\documentclass[a4paper,fleqn]{cas-dc}

\usepackage[numbers]{natbib}

\usepackage{amsmath, amssymb, physics, bbm, cancel}
\usepackage{wrapfig, multirow, makecell, float, hhline, booktabs, subcaption}
\usepackage{graphicx}
\usepackage{array,threeparttable}
\usepackage{enumitem, wasysym}
\usepackage[ruled,vlined]{algorithm2e}
\usepackage{xcolor}
\usepackage{listings}

\definecolor{AutoRECKeyword}{rgb}{0.00,0.50,0.00}
\definecolor{AutoRECNamespace}{rgb}{0.00,0.00,1.00}
\definecolor{AutoRECString}{rgb}{0.73,0.13,0.13}
\definecolor{AutoRECComment}{rgb}{0.24,0.48,0.48}
\definecolor{AutoRECNumber}{rgb}{0.40,0.40,0.40}

\lstdefinestyle{AutoRECBase}{
  basicstyle=\ttfamily\scriptsize,
  breaklines=true,
  frame=single,
  framerule=1pt,
  columns=fullflexible,
  keepspaces=true,
  showstringspaces=false,
  upquote=true
}

\lstdefinestyle{AutoRECPython}{
  style=AutoRECBase,
  language=Python,
  alsoletter={.},
  keywordstyle=\color{AutoRECKeyword}\bfseries,
  commentstyle=\color{AutoRECComment}\itshape,
  stringstyle=\color{AutoRECString},
  morekeywords=[2]{
    autorec.data_preparation,
    autorec.environment,
    autorec.agent,
    autorec.utils,
    autorec.factory
  },
  keywordstyle=[2]\color{AutoRECNamespace}\bfseries,
  morekeywords=[3]{
    EISDataPrep,
    EIS_ECM_Env,
    DDQN_ECM,
    set_global_seed,
    pipeline_builder
  },
  keywordstyle=[3]\color{AutoRECNamespace},
  literate=*
    {0}{{\color{AutoRECNumber}0}}1
    {1}{{\color{AutoRECNumber}1}}1
    {2}{{\color{AutoRECNumber}2}}1
    {3}{{\color{AutoRECNumber}3}}1
    {4}{{\color{AutoRECNumber}4}}1
    {5}{{\color{AutoRECNumber}5}}1
    {6}{{\color{AutoRECNumber}6}}1
    {7}{{\color{AutoRECNumber}7}}1
    {8}{{\color{AutoRECNumber}8}}1
    {9}{{\color{AutoRECNumber}9}}1
}

\lstdefinelanguage{AutoRECYAML}{
  sensitive=false,
  comment=[l]{\#},
  morestring=[b]",
  morestring=[b]',
  keywords={true,false,null,y,n},
  morekeywords=[2]{
    dataprep,path,mode,eis_features,
    environment,dataset_path,chromosome_HEAD_len,seed,
    agent,save_dir,random_seed,num_trials
  }
}

\lstdefinestyle{AutoRECYAMLStyle}{
  style=AutoRECBase,
  language=AutoRECYAML,
  keywordstyle=\color{AutoRECKeyword}\bfseries,
  keywordstyle=[2]\color{AutoRECKeyword}\bfseries,
  commentstyle=\color{AutoRECComment}\itshape,
  stringstyle=\color{AutoRECString}
}

\usepackage{hyperref}

\graphicspath{{./figures/}}

\newcommand{\chithresh}{\chi_{\text{thresh}}}
\newcommand{\Ravg}{R_{\text{avg}}}
\newcommand{\Rscore}{R_{\text{score}}}
\newcommand{\Rmag}{R_{\vert Z \vert}}
\newcommand{\Rphase}{R_{\phi}}

\SetKwInOut{Input}{Input}

\begin{document}
\let\WriteBookmarks\relax
\def\floatpagepagefraction{1}
\def\textpagefraction{.001}
\shorttitle{AutoREC: An RL Platform for ECM Generation}
\shortauthors{A. Jaberi et al.}

\title [mode = title]{AutoREC: A reinforcement learning platform for equivalent circuit model generation}

\author[1]{A. Jaberi}
\fnmark[1]
\credit{
    Writing--original draft,
    conceptualization,
    methodology,
    formal analysis,
    software
}

\author[2]{Y. Kurniawan}
\fnmark[1]
\credit{
    Writing--original draft,
    investigation,
    formal analysis,
    software
}

\author[1]{R. Black}
\credit{
    Conceptualization,
    investigation,
    resources,
    project administration
}

\author[1]{S. Mousavi M.}
\credit{
    Conceptualization,
    methodology,
    investigation
}

\author[3]{K. Verma}
\credit{
    Software
}

\author[1]{Z. Sadighi}
\credit{
    Validation
}

\author[4]{S. Miret}
\credit{
    Conceptualization,
    investigation
}

\author[2]{J. Hattrick-Simpers}
\cormark[1]
\ead{jason.hattrick.simpers@utoronto.ca}
\credit{
    Writing--review \& editing,
    conceptualization,
    methodology,
    investigation,
    formal analysis,
    validation,
    resources,
    funding acquisition,
    project administration,
    supervision
}

\affiliation[1]{
    organization={Clean Energy Innovation Research Centre, National Research Council Canada},
    city={Mississauga},
    postcode={L5K 1B4},
    state={ON},
    country={Canada}
}

\affiliation[2]{
    organization={Department of Materials Science and Engineering, University of Toronto},
    city={Toronto},
    postcode={M5S 3E4},
    state={ON},
    country={Canada}
}

\affiliation[3]{
    organization={Cheriton School of Computer Science, University of Waterloo},
    city={Waterloo},
    postcode={N2L 3G1},
    state={ON},
    country={Canada}
}

\affiliation[4]{
    organization={Lila Sciences},
    city={San Francisco},
    state={CA},
    country={United States of America}
}

\cortext[1]{Corresponding author}
\fntext[fn1]{These authors contributed equally to this work.}

\begin{abstract}
    This paper introduces AutoREC, an open-source Python platform for developing, training, and evaluating reinforcement learning (RL) agents that automatically generate equivalent circuit models (ECMs) from electrochemical impedance spectroscopy (EIS) data.
    Although ECMs are widely used to interpret EIS measurements, their identification typically relies on manual trial-and-error, requiring domain expertise and limiting scalability, particularly in autonomous experimental pipelines such as self-driving laboratories.
    In AutoREC, ECM generation is formulated as a Markov decision process in which an RL agent sequentially modifies a circuit topology based on the current state, available actions, and feedback from the resulting model.
    The platform supports an end-to-end workflow encompassing EIS preprocessing with selectable impedance representations, agent setup and training, ECM generation for new measurements, and visualization-based evaluation and analysis of agent decision-making.
    AutoREC implements a configurable Double Deep Q-Network (DDQN) agent with prioritized experience replay and a dedicated dead-loop mitigation strategy for navigating the complex circuit-generation action space efficiently.
    To demonstrate the platform, we trained a representative agent on 1,350 synthetic spectra from five reference circuit families, evaluated it on 150 held-out synthetic spectra, and applied it to 32 previously unseen experimental spectra from battery, corrosion, oxygen evolution reaction, and CO$_2$ reduction systems.
    These case studies illustrate the end-to-end capabilities of AutoREC while revealing challenges associated with experimental complexity and limited training-data coverage.
    The demonstrated agent serves as a reference implementation; AutoREC provides an extensible foundation through which users can develop and evaluate agents tailored to their specific electrochemical systems and research objectives.
\end{abstract}

\begin{graphicalabstract}
    \includegraphics[width=\textwidth]{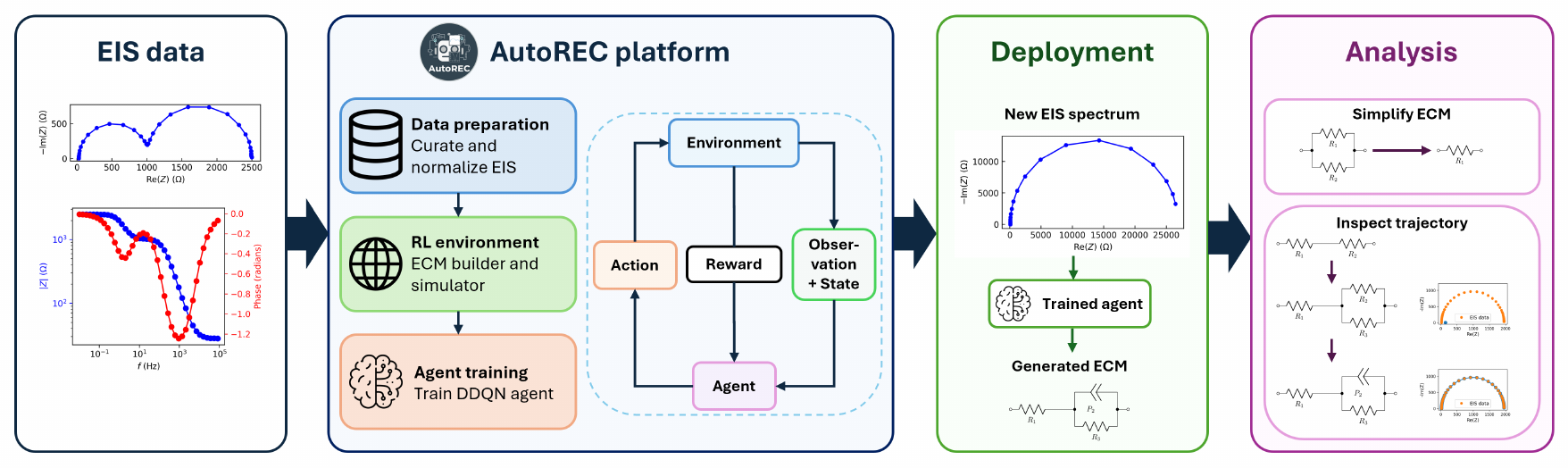}
\end{graphicalabstract}

\begin{highlights}
    \item AutoREC provides an open-source RL platform for automated ECM generation.
    \item The platform supports EIS data preparation, agent training, and deployment.
    \item Trained agents can generate ECMs for new EIS data without retraining.
    \item AutoREC provides tools to evaluate, simplify, and analyze generated ECMs.
    \item Synthetic and experimental EIS data demonstrate the complete AutoREC workflow.
\end{highlights}


\maketitle

\section{Introduction}
\label{sec:introduction}

Electrochemical systems underpin a wide range of important technologies, including batteries \cite{etacheri_challenges_2011}, fuel cells \cite{ormerod_solid_2003}, corrosion protection \cite{wei_study_2019}, and electrocatalytic energy conversion processes such as the oxygen evolution reaction (OER) \cite{tahir_electrocatalytic_2017} and CO$_2$ reduction \cite{liu_current_2020}.
The performance and durability of these systems are governed by coupled electrochemical processes such as charge transfer, double-layer formation, adsorption, and mass transport.
A detailed understanding of these processes is therefore essential for the rational design and optimization of electrochemical materials and devices.

Electrochemical impedance spectroscopy (EIS) is one of the most widely used non-destructive techniques for probing the dynamic behavior of electrochemical systems \cite{lasia_2014,modified_lazanas_electrochemical_2023}.
The measurement typically involves applying a sinusoidal signal (i.e., alternating current or voltage) to the cell across a broad range of frequencies and extracting the complex impedance from the measured current or voltage response.
This type of measurement enables the separation of physical processes according to their characteristic time scales.
For example, high-frequency responses are often associated with electrolyte resistance and interfacial charging, while lower-frequency behavior reflects charge-transfer kinetics, diffusion, and corrosion mechanisms.

A common framework for interpreting EIS data is the use of equivalent circuit models (ECMs), in which the electrochemical system is represented by a network of circuit elements such as resistors, capacitors, constant phase elements (CPEs), and diffusion-related components including Warburg or Gerischer elements.
To extract physically meaningful quantities from the impedance spectrum, an ECM is fitted to the EIS data to parameterize the contributions of different electrochemical processes.
The resulting parameters can be associated with quantities such as electrolyte resistance, charge-transfer resistance, double-layer capacitance, and mass transport coefficients.
In this way, ECMs provide a compact and interpretable representation of impedance spectra that is widely used for diagnosing electrochemical performance and identifying rate-limiting processes.

Traditionally, ECM analysis requires an expert to propose a circuit topology, fit its parameters, assess the resulting impedance response, and iteratively revise the model.
This process can be time-consuming and subjective, particularly when several circuit structures provide similarly good fits.
This manual workflow is also difficult to scale to the large volumes of data expected from automated and high-throughput experiments.
These limitations are especially relevant to self-driving laboratories (SDLs), which seek to integrate experiment planning, data acquisition, analysis, and decision-making within autonomous closed-loop workflows \cite{tom_self-driving_2024}.
In such settings, rapid and consistent interpretation of EIS measurements is necessary to prevent data analysis from becoming a bottleneck.

Recent work has therefore explored machine learning (ML) approaches to automate ECM identification from impedance data.
One line of research formulates ECM selection as a classification problem, in which a model predicts an appropriate circuit topology from a predefined library of candidate ECMs.
Methods investigated for this task include support vector machines \cite{zhu_equivalent_2019}, random forests \cite{chen_intelligent_2025}, gradient-boosted trees, convolutional neural networks \cite{schaeffer_machine_2023}, deep neural networks \cite{doonyapisut_analysis_2023,bongiorno_exploring_2022}, and large language models \cite{zhu_integrating_2025}.
While these approaches can achieve high accuracy within their training domain, they require labeled data for training or fine-tuning that may be costly to obtain and may inherit subjective biases from expert-selected circuit labels.
Furthermore, because their predictions depend on the candidate topologies and corresponding EIS distributions represented during training, they may not readily generalize to out-of-distribution EIS spectra or unseen circuit structures.

An alternative direction focuses on generative approaches that directly construct ECM structures.
A prominent example is gene expression programming (GEP), in which an ECM is represented as a chromosome and evolved through fitness-based selection and genetic operations \cite{cao_evolutionary_2003,arpaia_cultural_2009,ramos_gene_2013,van_haeverbeke_practical_2021}.
Subsequent work has incorporated Bayesian inference to improve parameter estimation, quantify uncertainty, and rank the candidate ECMs retained in the final population \cite{zhang_editors_2023}.
Despite their flexibility, GEP-based methods operate over a vast combinatorial space of possible circuit structures without strong guidance, effectively relying on stochastic exploration to identify suitable candidates.
As a result, the search process can be inefficient and sensitive to initialization, often requiring large populations and many evolutionary cycles to discover high-quality ECMs.
This lack of structured guidance limits the efficiency and reliability of evolutionary generative approaches.

From a modeling perspective, generating an ECM from an impedance spectrum can be viewed as a structured, sequential decision-making problem.
Starting from an initial, minimal circuit, the model is constructed through a sequence of decisions regarding which circuit elements to introduce, how they should be connected, and when construction should terminate.
Each modification affects the circuit's ability to reproduce the observed impedance response, its structural complexity, and the modifications available at subsequent steps.
This formulation naturally aligns with reinforcement learning (RL), where the current circuit and target impedance spectrum define the state, circuit modifications serve as actions, and fitting quality, structural validity, and model complexity inform the reward.
Unlike classification-based approaches, this formulation does not require a predefined library of circuit topologies or labeled EIS data; unlike evolutionary generative methods, it avoids relying primarily on unguided stochastic exploration.
Instead, the agent learns an adaptive circuit-construction policy that balances exploration and exploitation while directing the search toward promising regions of the ECM space.

A recent study demonstrated the feasibility of RL-based ECM discovery by representing circuit topology as an adjacency matrix and training a Q-learning agent to sequentially add connections among a predefined set of circuit elements \cite{chen_auto-eis_2025}.
While this work provided an important proof of concept for RL-based ECM construction, its methodology formulates RL as a spectrum-specific search, requiring the optimization to be repeated for each new EIS measurement.
In this work, we develop AutoREC (Autonomous Reinforcement ECM Composer), a Python library for developing and deploying RL agents for automated ECM generation from EIS data \cite{GitHub}.
Rather than optimizing an agent independently for each spectrum, AutoREC trains agents across a distribution of EIS spectra, with the target impedance spectrum incorporated into the agent state so that the learned circuit-construction policy can be reused across measurements.
To represent and modify circuit topology, AutoREC adopts a compact chromosome-based ECM representation adapted from GEP and formulates chromosome mutation as the agent's action space.
Agents are trained using reward signals that balance fitting accuracy and model parsimony.
Once trained, an agent can generate candidate ECMs for new measurements without requiring a separate RL optimization for each spectrum.
Beyond this specific agent formulation, AutoREC provides a flexible development environment through which users can train and evaluate agents tailored to their own datasets, circuit spaces, and EIS analysis tasks.

The remainder of this paper is organized as follows.
Section~\ref{sec:rl-formulation} presents the RL formulation for ECM generation implemented in AutoREC, including the state and action spaces, the reward formulation, and the training algorithm.
Section~\ref{sec:autorec} describes the AutoREC software platform, its design objectives and typical workflow, core modules, and additional functionalities provided for agent development, evaluation, and analysis.
Section~\ref{sec:case-study} presents a case study demonstrating the end-to-end use of AutoREC, including training a reference agent, evaluating it on synthetic and experimental EIS data, post-processing the generated circuits, and inspecting its circuit-construction trajectories.
The reference agent used in this demonstration is not presented as a definitive solution to automated ECM generation.
Rather, the primary contribution of this work is AutoREC as a software platform for developing, deploying, and analyzing RL agents for this task, with the longer-term goal of enabling scalable impedance analysis and supporting future integration into self-driving laboratory workflows.


\section{RL formulation for ECM generation}
\label{sec:rl-formulation}

\subsection{Problem formulation as a Markov Decision Process}
\label{sec:problem-formulation}

The task of generating an ECM for EIS data is formulated as a sequential decision-making problem and cast as a Markov Decision Process (MDP).
In this framework, an RL agent iteratively modifies the circuit topology, aiming to identify an ECM that achieves a high-quality fit to the target EIS.
Each modification of the circuit corresponds to an action taken by the agent, and improvements in the quality of fit are reflected through the reward signal.
To construct a valid MDP, the fundamental components of the RL environment, namely the state representation, action space, and reward function, are precisely defined as follows.

\subsection{State space}
\label{sec:state-space}

Since the electrical circuit is continuously modified by the agent's actions, the current circuit topology forms a core component of the state representation.
The circuit is represented using a linear chromosome, following the approach introduced by \citet{van_haeverbeke_practical_2021}.
In this formulation, the chromosome consists of a sequence of two types of symbols:
(i) terminals (circuit elements), including resistors, capacitors, inductors, and CPEs (represented in the chromosome as R, C, L, and P, respectively); and
(ii) functions (operators), namely + and /, denoting series and parallel connections, respectively.

\begin{figure*}[!hbt]
    \centering
    \includegraphics[width=0.9\textwidth]{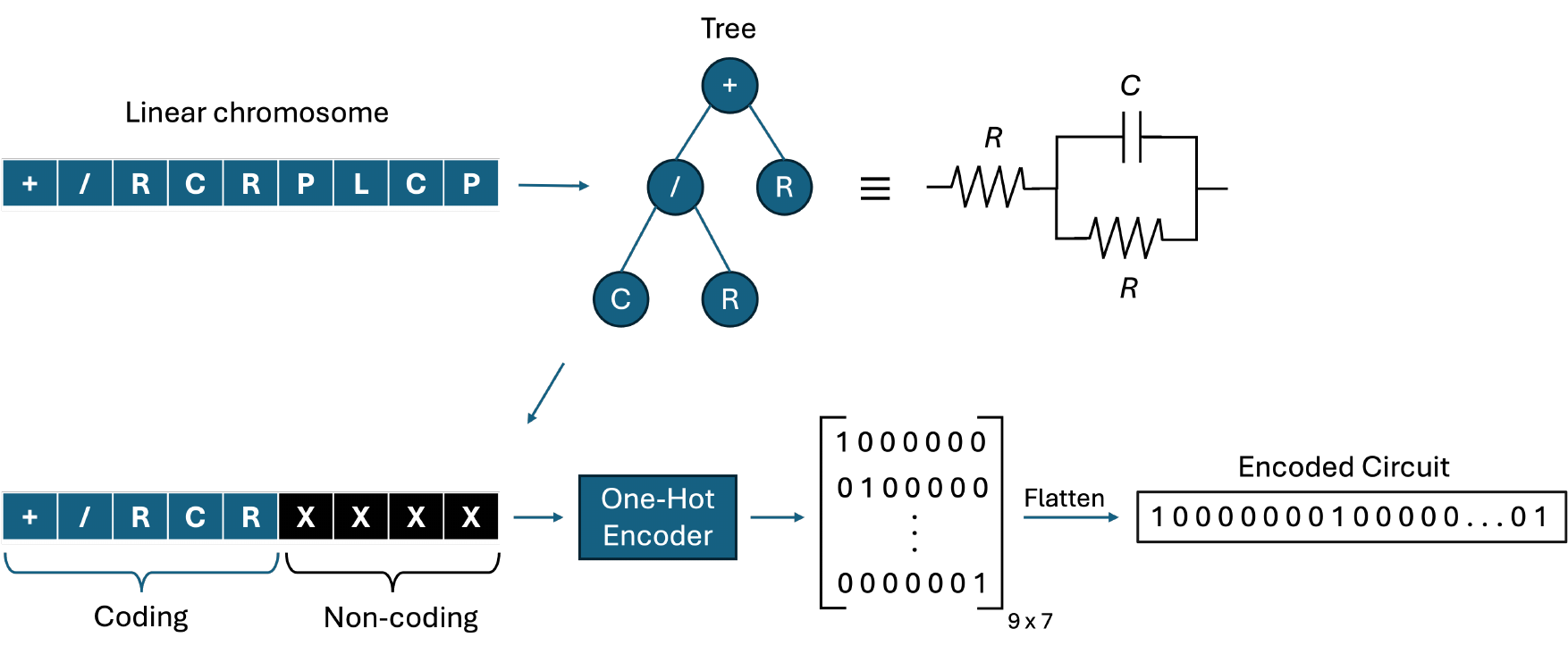}
    \caption{
	Pipeline for encoded circuit state construction using the linear chromosome representation of the circuit.
    }
    \label{fig:2-encoded-circuit}
\end{figure*}

The chromosome is divided into a head and a tail.
The head may contain both circuit elements and operators, whereas the tail is restricted to circuit elements only.
This linear chromosome can be deterministically transformed into a tree-based representation, which is then interpreted as an electrical circuit, as illustrated in Fig.~\ref{fig:2-encoded-circuit}.
During this transformation, some symbols in the linear chromosome may not contribute to the resulting circuit topology; these symbols constitute the non-coding part of the chromosome.
Although the non-coding part does not affect the physical circuit, it cannot be removed because the state representation must maintain a fixed length.
To address this, all symbols in the non-coding region are replaced with a dedicated non-coding token, denoted as X.
The resulting chromosome is then converted into a one-hot encoded vector and flattened to form the circuit part of the state.
An overview of this pipeline for encoded circuit construction is illustrated in Fig.~\ref{fig:2-encoded-circuit}.

To enable the agent to construct appropriate circuits for different EIS measurements, the observed EIS data are incorporated into the state representation.
Specifically, the EIS features are concatenated with the encoded circuit representation and jointly provided as input to the agent.
This formulation naturally aligns with the universal value function approximator (UVFA) framework \cite{schaul_universal_2015}, in which the goal is appended to the state of the function approximator.
In the present context, each EIS spectrum serves as a distinct goal within the UVFA formulation.

\begin{figure*}[!hbt]
    \centering
    \includegraphics[width=0.9\textwidth]{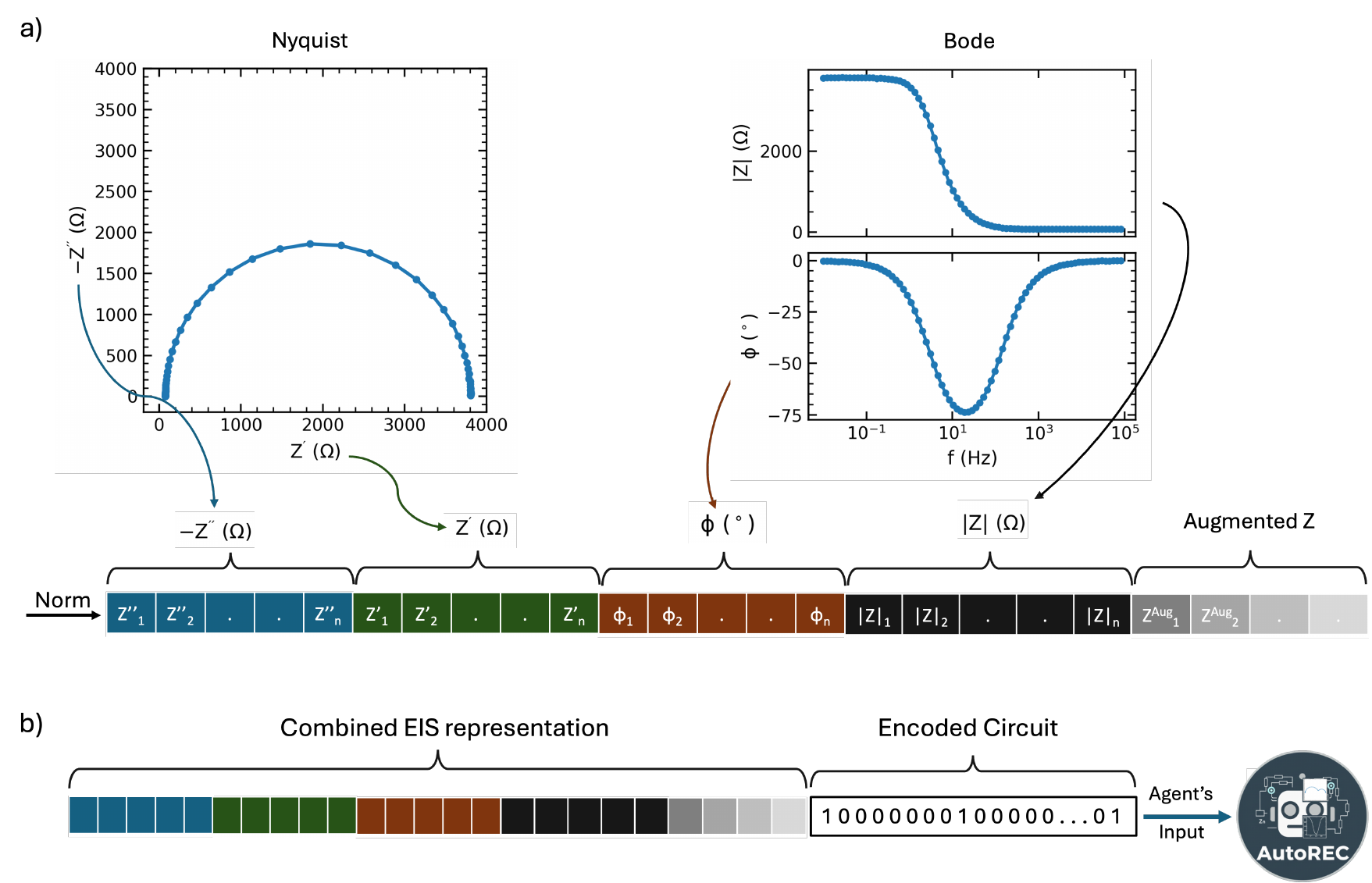}
    \caption{
	a) The pipeline for constructing the combined EIS representation and b) the final state input of the agent.
    }
    \label{fig:2-eis-representation}
\end{figure*}

Because EIS data are complex-valued, they may be expressed using multiple representations (e.g., Nyquist or Bode), each capturing complementary information.
To preserve this information, the real part, imaginary part, magnitude, and phase of each impedance point are extracted and normalized.
The EIS feature vector is then formed by concatenating some or all of these features; augmented variants, such as inverse or sign-inverted representations, may also be included to improve robustness.
An overview of this EIS state construction process is illustrated in Figure~\ref{fig:2-eis-representation}a. The final state input to the agent is obtained by concatenating the encoded circuit representation with the combined EIS feature vector (Fig. \ref{fig:2-eis-representation}b).

\subsection{Action space}
\label{sec:action-space}

\begin{figure*}[!hbt]
    \centering
    \includegraphics[width=0.9\textwidth]{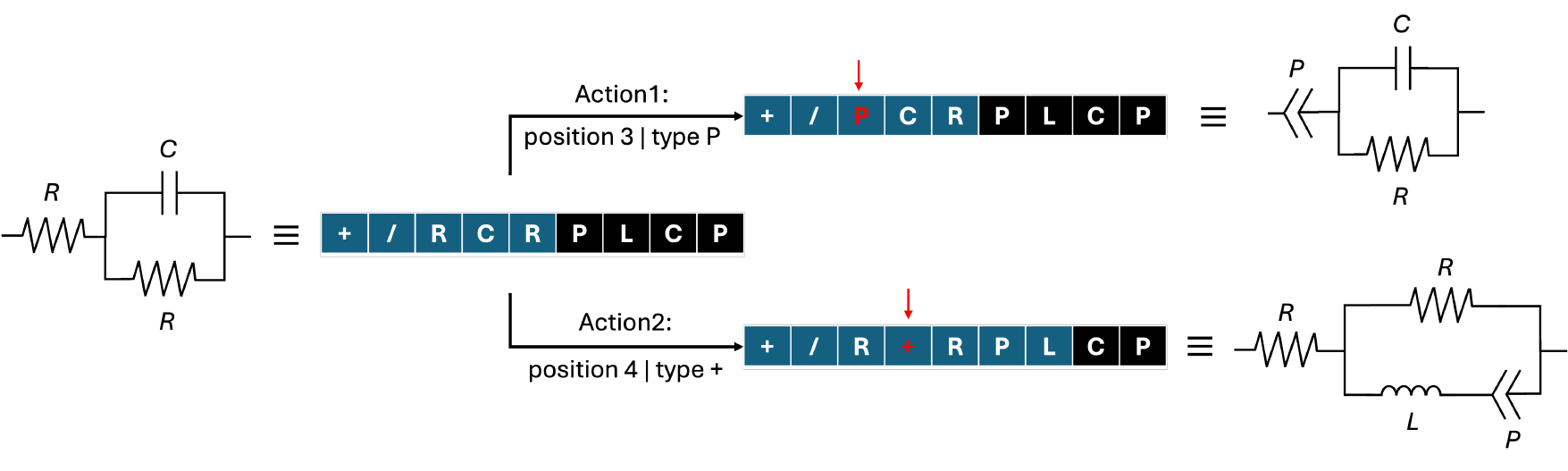}
    \caption{
	Two examples of agent's actions.
	The blue squares show the coding part, and the black squares show the non-coding part of the chromosome.
	The figure clearly shows that the action can change the length of the coding part, yielding circuits with different numbers of elements and varying complexity.
    }
    \label{fig:2-action-space}
\end{figure*}

The agent modifies the circuit topology by applying mutation operations to the linear chromosome representation.
Each action is defined by a pair consisting of the action position at which the mutation is applied and the action type, specifying the target symbol---either a circuit element or an operator.
Representative examples of such mutations are shown in Fig.~\ref{fig:2-action-space}.
This action formulation enables the agent to alter both the structure and the complexity of the circuit, as mutations may change the length of the coding region of the chromosome, thereby producing circuits with different numbers of elements and varying topological complexity.
As a result, each position along the chromosome can be mutated to one of $N$ discrete actions, corresponding to the total number of elements and operators.

However, a substantial fraction of the action space is invalid.
Certain mutations violate physical constraints, such as those that remove the ohmic resistance, which must be present in all physically meaningful ECMs.
Other mutations break the grammatical rules of the chromosome, for example by converting symbols in the tail region into operators.
In addition, some actions do not alter the circuit topology at all, either because a symbol is mutated to itself or because the mutation is applied to the non-coding region of the chromosome.
Whenever an invalid action is selected, the environment rejects the mutation and returns the agent to the previous state.

\subsection{Episode initialization and termination}
\label{sec:episode-initialization}

At the beginning of each episode, the environment is initialized with a specified circuit selected from one or more candidate initial circuits.
Starting from the selected topology, the agent progressively modifies the circuit by applying mutation actions throughout the episode.
This configurable initialization provides flexibility in defining the starting topology while allowing the agent to explore a wide range of circuit structures.

A terminal condition is required to determine when an episode should end and the environment should be reset.
An episode is terminated when the agent identifies a circuit that achieves a sufficiently high quality of fit to the target EIS data.
Rather than using a fixed termination threshold, which may be inappropriate given the variability in EIS characteristics and noise levels across datasets, a data-dependent criterion is employed.
Specifically, the termination threshold is set relative to the chi-squared ($\chi^2$) value obtained from fitting the EIS data with a linear Kramers--Kronig model \cite{schonleber_method_2014_edited}, using a configurable offset.
This criterion provides a spectrum-dependent reference for defining the required fit quality and a consistent, physically motivated termination condition across different EIS spectra.

\subsection{Reward system}
\label{sec:reward-system}

An ECM is defined by both its circuit topology and its component values.
The fitted component values are also needed to extract physical information about the system.
Therefore, each candidate topology is fitted to the target EIS data before it is evaluated.
The resulting fit quality is then used in the reward to guide the agent toward ECMs that accurately reproduce the target response.

After each action, the circuit generated by the agent is fitted to the target EIS data using the computationally efficient $\text{log-B}$ loss function \cite{jaberi_assessment_2026},
\begin{equation}
    \label{eq:loss}
    \begin{aligned}
      L_{\text{log-B}}(\mathbf{P}) =
      \sum_{i=1}^n \Bigg\{ &
                             \left[ \log_{10} \frac{
                             \vert Z^{\text{GT}}_i \vert
                             }{
                             \vert Z^{\text{ECM}}(f_i, \mathbf{P}) \vert
                             } \right]^2 \\
      & + \left[\theta^{\text{GT}}_i - \theta^{\text{ECM}}(f_i, \mathbf{P}) \right]^2
      \Bigg\},
    \end{aligned}
\end{equation}
where the superscripts GT and ECM denote the ground truth and ECM prediction, respectively; $\vert Z \vert$ and $\theta$ denote the magnitude (in $\Omega$) and phase (in radians) of the complex impedance; $f$ denotes the signal frequency (in Hz); and $\mathbf{P}$ denotes the ECM component values.
The reward is then assigned based on the resulting fit quality, necessitating efficient evaluation of the loss.
Actions that produce invalid circuits receive a large negative reward (i.e., penalty), discouraging the agent from exploring physically or grammatically infeasible regions of the action space.
Conversely, when a terminal state is reached, defined by achieving a chi-squared value below the data-dependent threshold $\chithresh^2$, a large positive reward is assigned.
The terminal reward is determined using a combination of goodness-of-fit metrics to robustly assess model quality.
Specifically, both the $\chi^2$ value and the coefficient of determination ($R^2$) are considered, where the latter is computed as the average of $R^2$ values for different EIS representations.
This $\Ravg^2$ includes the Nyquist representation and the magnitude and phase components of the Bode representation, yielding the metrics $\Rscore^2$, $\Rmag^2$, and $\Rphase^2$.
This multi-metric formulation is motivated by prior work showing that no single metric is sufficient to fully characterize the quality of EIS fits \cite{jaberi_assessment_2026}.

To discourage overly complex circuit topologies, additional collective penalties $P$ are incorporated into the terminal reward.
These penalties account for both the total number of circuit elements and the depth of nested parallel branches, with increasing structural complexity penalized to promote simpler and more interpretable ECMs.
Because the large positive and negative rewards are received only upon successful termination or the generation of an invalid circuit, they provide limited feedback for valid intermediate actions.
Small positive and negative rewards are therefore assigned during an episode to reflect incremental improvements or degradations in fit quality, thereby reducing reward sparsity and guiding the agent toward progressively improved circuit models.

Together, these components define a hierarchical reward structure that balances fitting accuracy, physical validity, and model simplicity.
The reward-function hierarchy is defined as follows:

\begin{equation}
    \label{eq:reward}
    \begin{aligned}
      r\left(s,a\mid\text{EIS}\right)
      &=
        \begin{cases}
          -0.5,
          & \text{if action is invalid}, \\
          \mathcal{R}\left(\chi^2,\Ravg^2\right),
          & \text{if } \chi^2 < \chithresh^2, \\
          +0.01,
          & \text{if } \min\mathcal{S}_{R^2} > 0.5, \\
          -0.01,
          & \text{if } \min\mathcal{S}_{R^2} > 0, \\
          -0.02,
          & \text{otherwise},
        \end{cases}
      \\
      \mathcal{R}\left(\chi^2,\Ravg^2\right)
      &= \log_{10}\left(\frac{1}{\chi^2}\right)
        + \log_{10}\left(\frac{1}{1-\Ravg^2}\right) + P, \\
      \mathcal{S}_{R^2}
      &= \left\{\Rscore^2, \Rmag^2, \Rphase^2\right\}.
    \end{aligned}
\end{equation}

\subsection{DDQN training algorithm}
\label{sec:training-algorithm}

Given the discrete action space, we employ a Double Deep Q-Network (DDQN) algorithm \cite{hasselt_deep_2016,mnih_playing_2013,mnih_human-level_2015} to maximize the expected cumulative discounted reward through sequential circuit modifications.
DDQN extends deep Q-learning, a model-free RL method that learns an approximation to the optimal action-value function (Q-function), from which an optimal policy can be obtained \cite{Sutton1998-rs}.
The optimal action-value function $Q^*$ represents the maximum expected discounted return obtainable after taking action $a_t$ in state $s_t$ and acting optimally thereafter.
For a deterministic transition, it satisfies the Bellman optimality equation
\begin{equation}
    \label{eq:q-function}
    Q^*\left(s_t, a_t \mid \text{EIS}\right) = r_t + \gamma \max_a Q^*\left(s_{t+1}, a \mid \text{EIS}\right),
\end{equation}
where $r_t$ is the reward at time step $t$, $\gamma$ is the discount factor, and the EIS spectrum conditions the action-value function through the UVFA formulation.

In the DDQN framework, the Q-function is approximated using a neural network.
Two networks with the same architecture are employed: an online network with parameters $\theta$, which is updated via gradient descent, and a target network with parameters $\theta^-$, whose parameters are periodically updated by copying those of the online network.
In practice, DDQN constructs an approximation to the Bellman optimality target in Eq.~\eqref{eq:q-function} using the online and target networks for action selection and evaluation, respectively.
For a transition $(s_t,a_t,r_t,s_{t+1})$, the online network selects the action that maximizes the estimated Q-value at the next state, while the target network evaluates the value of that action.
The corresponding DDQN target is
\begin{equation}
    \label{eq:ddqn-target}
    \begin{aligned}
      a^*_{t+1} &= \underset{a}{\arg\max}\, Q(s_{t+1},a\mid\text{EIS};\theta), \\
      y_t &= r_t + \gamma Q(s_{t+1},a^*_{t+1}\mid\text{EIS};\theta^-).
    \end{aligned}
\end{equation}
This decoupling of action selection and action evaluation by two networks mitigates overestimation bias and improves the stability of training \cite{hasselt_deep_2016}.

To further enhance learning efficiency in the presence of a large and complex state space, prioritized experience replay (PER) \cite{schaul_prioritized_2016} is used to sample transitions from the replay buffer according to their priorities rather than uniformly.
The priority of each transition is determined from its temporal-difference error, which measures the discrepancy between the current Q-value estimate $Q(s_t,a_t\mid\text{EIS};\theta)$ and the corresponding DDQN target $y_t$, allowing transitions with larger errors to be sampled more frequently.
The importance-sampling correction factor $\beta$ used in PER is gradually increased over episodes toward a specified maximum to reduce sampling bias as training progresses.
Exploration is handled using an $\epsilon$-greedy strategy, in which $\epsilon$ is simultaneously decreased over the same episodes until a predefined minimum value is reached, gradually shifting the agent from exploration toward exploitation of learned policy.

To address the challenge posed by the large proportion of invalid actions in the action space \cite{huang_closer_2022_modified,chen_auto-eis_2025}, we introduce a task-specific dead-loop mitigation heuristic during training.
Rather than modifying the underlying RL algorithm, this procedure is intended to stabilize learning in an environment where many actions leave the circuit topology unchanged.
In the early stages of learning, the agent may overestimate the value of invalid actions, leading to repeated selection of such actions.
Because invalid actions do not alter the circuit topology, the agent may remain in the same state for an entire episode, resulting in highly redundant transitions being stored in the replay buffer and reducing the diversity of informative experiences.

To prevent this behavior, the agent monitors both the number of consecutive identical actions and the number of times the same next state is encountered within an episode.
If either count exceeds a predefined threshold, the episode is identified as being trapped in a dead loop.
For the subsequent action, the agent is temporarily restricted to selecting only valid actions.
By retaining invalid actions in the action space while incorporating this dead-loop detection strategy, the agent is encouraged to learn to recognize invalid actions while maintaining sufficient diversity in the replay buffer to support effective learning.
The pseudocode for this dead-loop detection and mitigation is provided in Algorithm~\ref{alg:dead-loop}.

\begin{algorithm}[!hbt]
    \caption{Stepwise Dead-Loop Detection and Mitigation}
    \label{alg:dead-loop}

    \Input{
	Thresholds $N_{\text{action}}$, $N_{\text{state}}$
	\tcp{Thresholds for the number of consecutive repeated actions and states, respectively}
    }

    \For{episode = $1 : M$}{
	Reset environment and observe initial state $s_0$\;
	Initialize episode history $\mathcal{H} \leftarrow [\ ]$\;
	Set $\text{DL}_{\text{flag}} \leftarrow \text{False}$\;
	Set $n_{\text{invalid}} \leftarrow 0$\;
	Set $n_{\text{state}} \leftarrow 0$\;
	\For{t = $1 : t_{\text{max}}$}{
	    \If{$\text{DL}_{\text{flag}} = \text{True}$}{
		Restrict action space to valid actions only\;
	    }
	    \Else{
		Use full action space\;
	    }
	    Select action $\left( \text{type}_t, \text{pos}_t \right)$ using $\epsilon$-greedy\;
	    Execute action, observe reward $r_t$, next state $s_{t+1}$, and validity $v_t$\;
	    Append transition record (including $\left( \text{type}_t, \text{pos}_t \right)$ and $s_{t+1})$ to $\mathcal{H}$\;
	    \tcp{Consecutive identical invalid-action check}
	    \If{$v_t = 0$ \textbf{and} $t > 0$}{
		\If{$\text{type}_t = \text{type}_{t-1}$ \textbf{and} $\text{pos}_t = \text{pos}_{t-1}$}{
		    $n_{\text{invalid}} \leftarrow n_{\text{invalid}} + 1$\;
		}
		$n_{\text{invalid}} \leftarrow 0$\;
	    }
	    \Else{
		$n_{\text{invalid}} \leftarrow 0$\;
	    }
	    \tcp{State revisitation count over entire episode history}
	    $n_{\text{state}} \leftarrow \sum_{h \in \mathcal{H}} \mathbf{1}\!\left[h.\text{new\_state} = s_{t+1}\right]$\;
	    \tcp{Dead-loop detection (stepwise)}
	    \If{$n_{\text{invalid}} \geq N_{\text{action}}$ \textbf{or} $n_{\text{state}} \geq N_{\text{state}}$}{
		$\text{DL}_{\text{flag}} \leftarrow \text{True}$\;
	    }
	    $s_t \leftarrow s_{t+1}$\;
	}
    }
\end{algorithm}


\section{AutoREC platform}
\label{sec:autorec}

Translating the RL formulation described in Sec.~\ref{sec:rl-formulation} into a practical ECM-generation workflow requires coordinated tools for EIS data preparation, environment construction, agent training, deployment, and analysis.
We therefore developed AutoREC (Autonomous Reinforcement ECM Composer), an open-source Python platform that provides reusable infrastructure for these tasks.
By separating these capabilities from any individual agent implementation, AutoREC reduces the effort required to develop and evaluate new agents while enabling trained policies to be applied consistently to newly acquired measurements.
This reusable workflow also provides a foundation for integrating automated EIS analysis into self-driving laboratory pipelines.
AutoREC is publicly available on GitHub \cite{GitHub}, and the following sections describe version 0.2.0, the latest release at the time of writing.

\subsection{Design objective and workflow}
\label{sec:design-objective-workflow}

AutoREC was designed to provide an integrated, reproducible, and extensible workflow for automated ECM generation from EIS measurements.
The platform reduces the manual effort required to prepare EIS datasets, configure reinforcement-learning experiments, train agents, and apply trained models to new measurements.
The package is organized around three primary components, implemented as the \texttt{EISDataPrep}, \texttt{EIS\_ECM\_Env}, and \texttt{DDQN\_ECM} classes.
Its modular, class-based organization allows individual components of the workflow to be configured independently and further extended without modifying the core package.
Because these components are implemented as classes, users can define custom subclasses that inherit the existing functionality and override only the methods needed for a particular modification, such as the reward calculation or agent behavior.
This design enables users to customize specific aspects of AutoREC without having to reimplement the remaining functionality of the corresponding class.

\begin{figure}[pos=!htbp]
    \centering
    \includegraphics[width=0.45\textwidth]{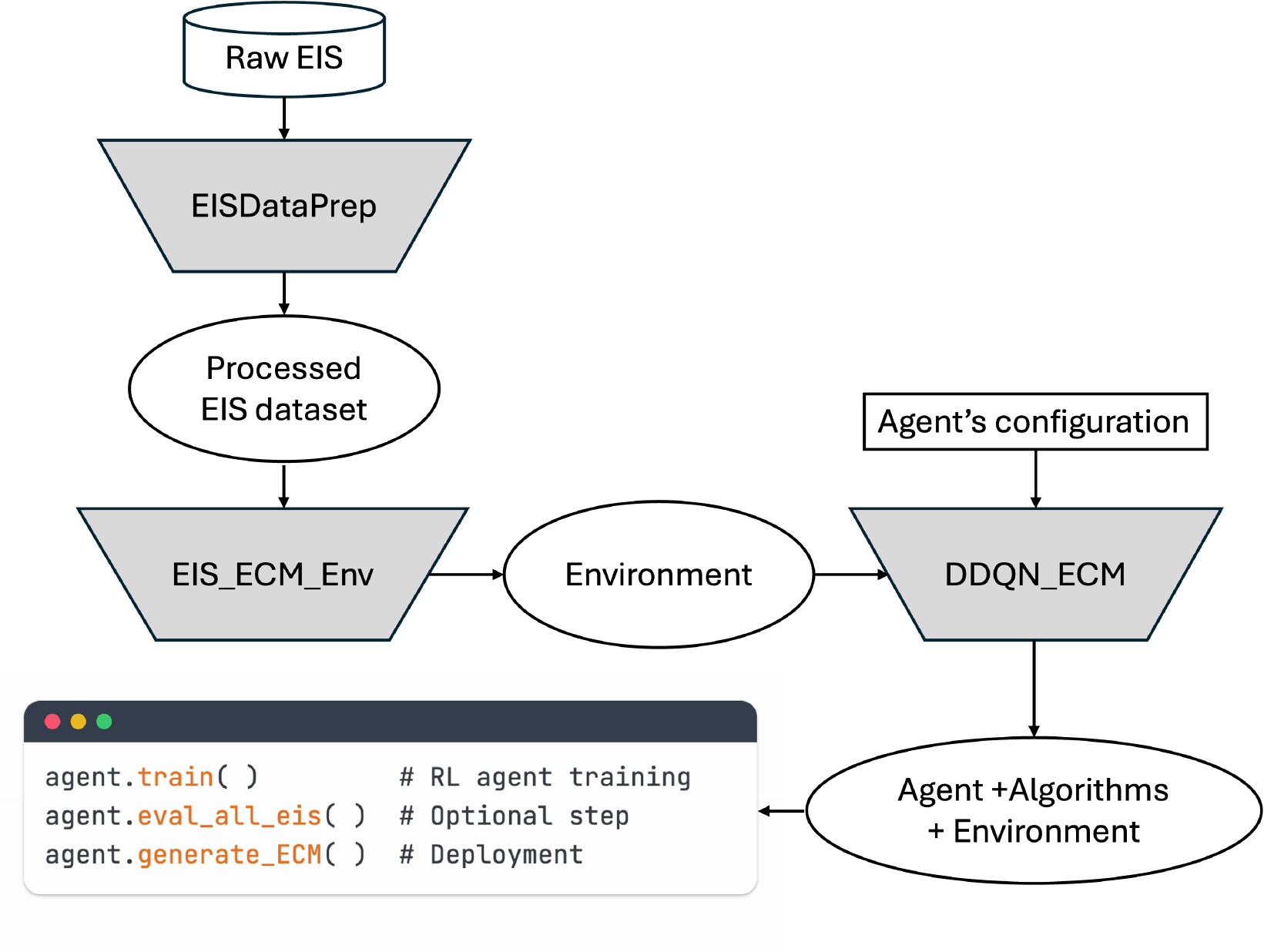}
    \caption{
        Workflow for using the AutoREC package.
        The three shaded trapezoids represent the three primary components of AutoREC: \texttt{EISDataPrep}, \texttt{EIS\_ECM\_Env}, and \texttt{DDQN\_ECM}.
    }
    \label{fig:3-autorec}
\end{figure}

The overall AutoREC workflow is illustrated in Fig.~\ref{fig:3-autorec}.
The workflow begins with the preparation and validation of the EIS dataset.
The processed data and parameter settings are then used to initialize the training environments.
AutoREC can also load a separate evaluation dataset, enabling periodic performance assessment during training as well as final evaluation after training is completed.
After the agent and its training parameters have been configured, the training procedure is executed.
The trained model can subsequently be used to generate ECMs for individual spectra or selected subsets of a dataset.
The resulting circuits, fitted parameters, goodness-of-fit metrics, and action histories are retained for analysis and optional post-processing.

AutoREC supports both direct Python-based initialization and YAML-based configuration.
Direct initialization provides greater flexibility for interactive development and methodological modifications, including the use of customized classes derived from AutoREC components.
YAML configuration files provide a concise and reproducible specification of the standard workflow and the settings associated with each training run.

\subsection{Data preparation}
\label{sec:data-preparation}

The \texttt{EISDataPrep} class transforms EIS data into the dataset format required by the RL environment.
It can either process raw EIS data or load an existing processed dataset.
Raw data are supplied as CSV files containing the frequency and the real and imaginary parts of the impedance specified by the \texttt{freq}, \texttt{Z\_real}, and \texttt{Z\_imag} columns, respectively.
During preprocessing, the class applies a linear Kramers--Kronig analysis to identify and remove inconsistent data points and uses cubic-spline interpolation to map the spectra onto a common frequency grid.
The interpolation ensures a consistent input dimension across spectra, which is required by the neural-network agent.

The EIS component of the agent input is then generated by normalizing the interpolated spectra and concatenating a user-selected combination of impedance features.
AutoREC supports representations based on the real and imaginary parts of impedance, phase, magnitude, and their sign-reversed counterparts.
By default, the representation concatenates the imaginary impedance, phase, magnitude, and negative phase.
The resulting concatenated representation is stored as \texttt{flatten\_Z} in the processed dataset.
Because feature selection is performed during preprocessing, alternative input representations can be investigated without modifying either the environment or the agent.

For each spectrum, the class also calculates the data-dependent fitting thresholds described in Sec.~\ref{sec:reward-system}.
These thresholds are derived from the linear Kramers--Kronig fit and stored alongside the processed dataset.
The following code demonstrates how to process the raw EIS data and save the processed dataset as a Python pickle file:
\begin{center}
    \begin{minipage}{0.475\textwidth}
        \centering Snippet: Data preparation
        \vspace{-6pt}
        \begin{lstlisting}[style=AutoRECPython]
from autorec.data_preparation import EISDataPrep

# set up EIS data processing
data_prep_raw = EISDataPrep(
    path="EIS_raw_demo",  # Folder containing subfolders with CSVs
    mode="process",  # Process raw files
    eis_features=["ImZ", "phi", "mag", "nphi"],  # Features to use
)
# processed raw data stored in raw_folder
dataset_raw = data_prep_raw.load()
# save the processed dataset
dataset_raw.save("data/training.pkl")
        \end{lstlisting}
    \end{minipage}
\end{center}

\subsection{RL environment}
\label{sec:rl-environment}

The \texttt{EIS\_ECM\_Env} class defines the RL environment for agent interaction.
At the beginning of each training episode, the environment samples an EIS spectrum from the processed dataset and selects an initial ECM state from the provided list of initial states.
By default, this list contains ECMs consisting of two, three, and four resistors connected in series.
Following each valid action, the environment updates the circuit representation, converts the expressed chromosome into a conventional circuit description, and fits the candidate ECM to the selected EIS spectrum.
The fitting procedure estimates the circuit parameters and calculates the corresponding predicted impedance.
The environment then evaluates the candidate using the fitting metrics and reward formulation introduced in Sec.~\ref{sec:reward-system}.
Its main output includes the updated observation, reward, and termination status.

The environment also identifies invalid circuit modifications before parameter fitting.
Invalid transitions are assigned the corresponding penalty without performing the comparatively expensive fitting calculation.
This validation ensures that the agent interacts only with circuit topologies permitted by the representation defined in Sec.~\ref{sec:rl-formulation}.

The maximum complexity of the circuits accessible to the agent is controlled by the chromosome head length.
The environment accepts user-defined initial states, random seeds, and head length.
These parameters make it possible to adapt the search space and computational requirements to the ECM families and dataset under investigation.

\subsection{DDQN agent, training, and deployment}
\label{sec:ddqn-agent}

The \texttt{DDQN\_ECM} class implements agent initialization, training, model persistence, and inference.
By default, the agent uses a fully connected neural network implemented in TensorFlow \cite{tensorflow2015-whitepaper}, with two hidden layers of 40 neurons each and rectified-linear-unit (ReLU) activation functions.
The network architecture is configurable, allowing users to specify the number and size of hidden layers as well as the activation function.
The output layer contains one value for each action defined by the associated environment.

During training, the agent repeatedly samples spectra through the environment and stores the resulting transitions in a fixed-capacity replay buffer.
AutoREC supports prioritized experience replay and periodically updates the target network.
The class also implements the dead-loop mitigation strategy described in Sec.~\ref{sec:training-algorithm}.
The exploration schedule, replay-buffer parameters, learning rate, batch size, update frequencies, discount factor, and number of training trials are configurable.
Intermediate neural-network models can be saved at a user-defined frequency, and the final trained model is stored by default.

During deployment, exploration is disabled and the trained agent selects actions greedily.
AutoREC can generate an ECM for an individual spectrum, a selected subset of a dataset, or every spectrum in the active evaluation environment.
The output for each spectrum includes the best circuit encountered, its fitted parameters and metrics, whether the terminal criterion was satisfied, the number of actions performed, and the complete sequence of circuit modifications.
When labeled spectra are available, the generated and reference ECMs can subsequently be compared.

When applying a previously trained agent to new EIS spectra, the data-preparation and environment settings must remain compatible with those used during training.
This includes the EIS representation, the number of frequency points in the processed spectrum, the allowed circuit elements, and the chromosome head length.
Changing any of these settings would alter the EIS input, state space, and/or action space expected by the trained agent and thus requires training a new policy.
The raw measurements do not need to contain the same number of frequency points, because \texttt{EISDataPrep} interpolates each spectrum onto the frequency grid required by the trained agent.
These and other workflow settings can be specified through the configuration interface described in Sec.~\ref{sec:configuration-development}.

\subsection{Post-processing and analysis}
\label{sec:postprocessing-analysis}

AutoREC provides optional post-processing and analysis utilities for examining agent performance and circuit-generation trajectories.
Generated ECMs may contain elements that are electrically reducible, such as multiple resistors, capacitors, or inductors connected entirely in series or parallel.
AutoREC therefore provides a post-processing utility that simplifies these structures and recalculates their equivalent component values.
A constant-phase element can additionally be converted to an ideal resistive or capacitive element when its fitted exponent satisfies user-defined criteria.
This optional simplification is performed after inference and does not modify the agent's original output or training process.

When reference ECMs are available, structural agreement can be evaluated before and after simplification.
This distinction is important because two circuit descriptions may be electrically equivalent even when their unsimplified symbolic representations differ.
AutoREC also retains the action history generated during evaluation, which can be used to examine individual circuit-generation trajectories.
When GIF generation is enabled, this history is used to produce an animated GIF showing the stepwise evolution of the ECM.

\subsection{Configuration and development}
\label{sec:configuration-development}

AutoREC experiments can be configured either by passing arguments directly to the three primary classes or by using YAML configuration files.
The three primary classes can be instantiated directly when individual settings must be specified or modified programmatically.
The following example constructs separate training and evaluation environments, trains the agent, and applies the trained policy to the evaluation dataset:
\begin{center}
    \begin{minipage}{0.475\textwidth}
        \centering Snippet: Agent training and evaluation
        \vspace{-6pt}
        \begin{lstlisting}[style=AutoRECPython]
from autorec.data_preparation import EISDataPrep
from autorec.environment import EIS_ECM_Env
from autorec.agent import DDQN_ECM
from autorec.utils import set_global_seed

set_global_seed(42, deterministic_ops=True)

# load processed dataset
train_data = EISDataPrep("data.training.pkl", mode="load").load()
eval_data = EISDataPrep("data.evaluation.pkl", mode="load").load()
# set up RL environment
train_env = EIS_ECM_Env(
    dataset=train_data, seed=42, chromosome_HEAD_len=10
)
eval_env = EIS_ECM_Env(
    dataset=eval_data, seed=42, chromosome_HEAD_len=10
)
# set up agent
agent = DDQN_ECM(
    training_env=train_env, eval_env=eval_env,
    num_trials=10000, save_dir="results/run_01",
    random_seed=42,
)
# train the agent
agent.train()
# generate circuits and evaluate them
results = agent.eval_all_eis(use_eval_env=True)
        \end{lstlisting}
    \end{minipage}
\end{center}

Alternatively, the complete workflow configuration can be specified in a single YAML file, providing a concise and reproducible definition of the data-preparation, environment, and agent settings.
The factory interface uses this configuration to construct the corresponding datasets, RL environments, and agent, after which training and evaluation can be invoked directly through the agent methods:
\begin{center}
    \begin{minipage}{0.475\textwidth}
        \centering Snippet: YAML configuration
        \vspace{-6pt}
        \begin{lstlisting}[style=AutoRECYAMLStyle]
# config.yaml
dataprep:
  path: EIS_raw_demo
  mode: process
  eis_features:
    - ImZ
    - phi
    - mag
    - nphi
environment:
    dataset_path: ./data/training.pkl
    chromosome_HEAD_len: 10
    seed: 42
agent:
    save_dir: ./results/run_01
    random_seed: 42
    num_trials: 10000
        \end{lstlisting}
    \end{minipage}
\end{center}

\begin{center}
    \begin{minipage}{0.475\textwidth}
        \centering Snippet: Agent training and evaluation using YAML configuration
        \vspace{-6pt}
        \begin{lstlisting}[style=AutoRECPython]
from autorec.factory import pipeline_builder

# set up dataset, RL environments and agent from YAML file
dataset, eval_dataset, env, eval_env, agent = pipeline_builder(
    "config.yaml"
)
# train agent
agent.train()
# generate circuits and evaluate them
results = agent.eval_all_eis(use_eval_env=True)
        \end{lstlisting}
    \end{minipage}
\end{center}


\section{Case study: Training and evaluating a reference agent}
\label{sec:case-study}

To demonstrate the capabilities of AutoREC, we present a case study in which a reference DDQN agent is configured and trained using the platform and subsequently evaluated on synthetic and experimental EIS data.
The case study illustrates an end-to-end workflow encompassing agent training, batch evaluation, postprocessing and analysis of generated ECMs, and application to previously unseen experimental spectra.
We also examine the agent's circuit-construction trajectories to illustrate how the evaluation records retained by AutoREC can be used to investigate recurring strategies and gain insight into the learned policy.
To provide an external point of reference, we additionally compare the reference agent with AutoEIS, an automated ECM identification method that uses GEP to generate candidate circuits \cite{zhang_editors_2023}.
This comparison is intended to contextualize the reference agent used in the case study rather than to establish agent performance as the primary focus of the work.

\subsection{Agent training}
\label{sec:agent-training}

\paragraph{Reference agent architecture}
For this case study, we configured and trained a reference DDQN agent using the default neural-network architecture and EIS representation described in Sec.~\ref{sec:ddqn-agent} and Sec.~\ref{sec:data-preparation}, respectively.
The network receives an observation that combines the current ECM representation with the target EIS representation, as described in Sec.~\ref{sec:state-space}, and outputs an estimated action value for each circuit-modification action.
The circuit-element space was restricted to R, L, and P.
Ideal capacitors were excluded because they can be represented as a special case of a CPE.
This configuration serves as a reference implementation for demonstrating the agent-training and deployment workflows supported by AutoREC, rather than as an exhaustively optimized agent architecture.

\begin{figure*}[!hbt]
    \centering
    \includegraphics[width=0.9\textwidth]{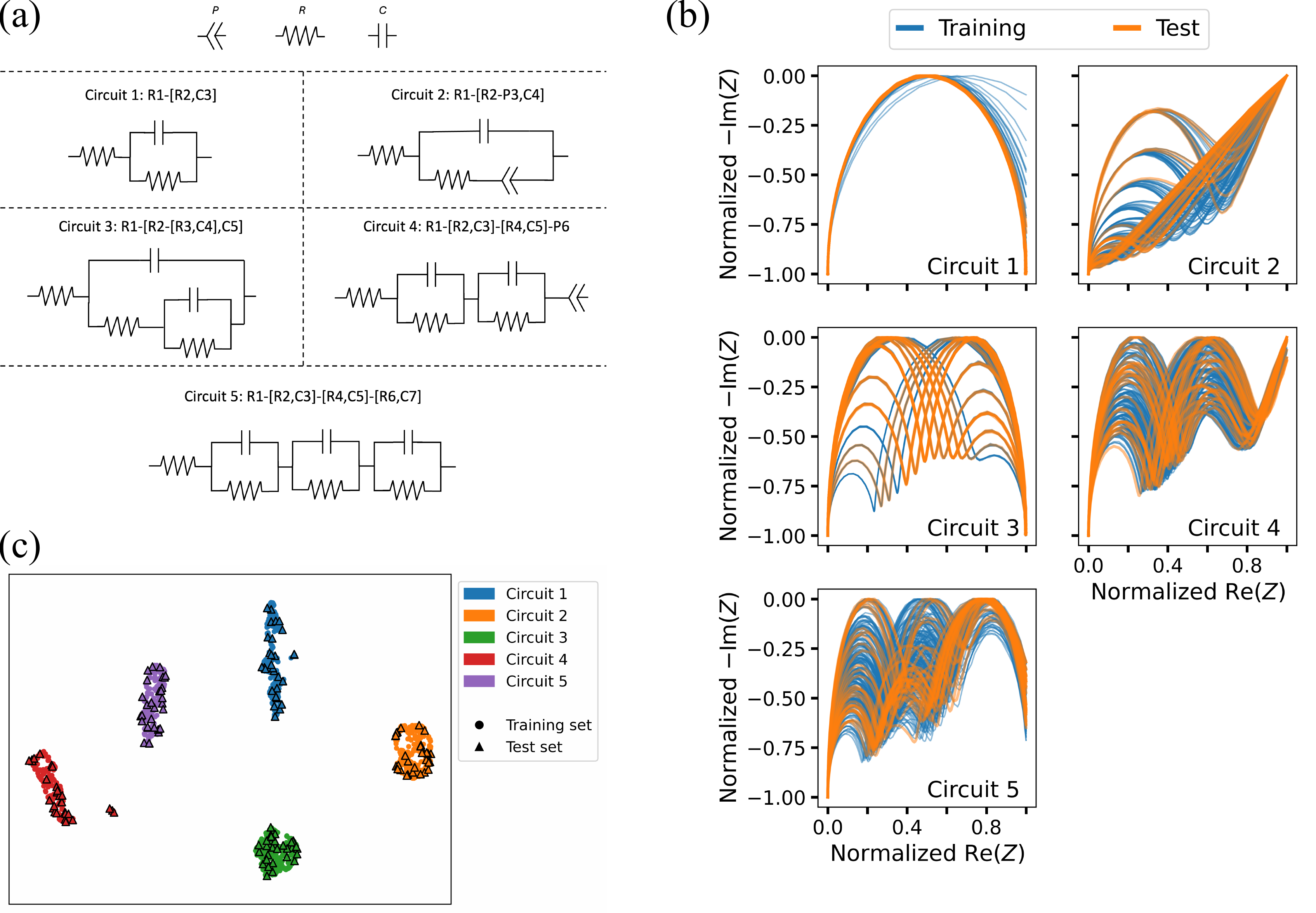}
    \caption{
	Representation of the EIS dataset used for demonstration.
	(a) Schematic diagrams of the five ECMs used to generate the dataset.
	(b) The corresponding Nyquist plot for each circuit.
	(c) UMAP projection of the dataset using the combined EIS representation, showing five well-separated clusters, each associated with a distinct ECM.
    }
    \label{fig:4-dataset}
\end{figure*}

\paragraph{Dataset}
To train the reference agent under controlled and reproducible conditions, we generated a synthetic EIS dataset following the procedure described by \citet{makogon_is_2024}.
The dataset contains spectra generated from five reference ECM topologies, shown in Fig.~\ref{fig:4-dataset}a.
These topologies were selected to represent several common circuit structures with varying levels of complexity; however, they were not intended to capture the full diversity of experimental EIS responses.
For each topology, 300 spectra were generated and divided into training and test sets using a $90:10$ ratio, yielding 1,350 training spectra and 150 test spectra.
The test set was reserved for evaluation of the trained agent.

The training and test spectra are visualized using Nyquist plots in Fig.~\ref{fig:4-dataset}b and a lower-dimensional embedding obtained using uniform manifold approximation and projection (UMAP) in Fig.~\ref{fig:4-dataset}c \cite{mcinnes_umap_2020}.
UMAP was fitted to the selected EIS representations of the training spectra, after which the test spectra were projected into the learned embedding.
The five ECM families form distinct clusters in the UMAP embedding, while the training and test spectra overlap within each cluster.
This result provides a qualitative view of the separation among the synthetic circuit families and indicates that the held-out spectra occupy regions of the embedding similar to those occupied by their corresponding training data.

\paragraph{Hyperparameter tuning}
The hyperparameters of the reference agent were optimized using the Optuna Python package \cite{akiba_optuna_2019} with the Tree-structured Parzen Estimator (TPE) sampler \cite{watanabe_tree-structured_2025}.
The optimization study evaluated 50 hyperparameter configurations, sampled in 10 batches of five trials from the search space defined in Table~\ref{tab:hyperparameters-list}.
The study was initialized with a hyperparameter configuration that had previously produced reasonable performance in preliminary experiments, after which the remaining configurations were selected by the TPE sampler.
Each configuration was evaluated after a training run of 10,000 episodes.

The optimization objective was to maximize a composite score defined as the mean of two quantities:
(i) the normalized average reward over the final 100 training episodes; and
(ii) the \emph{success rate} on training set, defined as the fraction of episodes in which the agent terminated successfully and generated a good-fitting ECM that satisfied the data-dependent $\chi^2_{\mathrm{thresh}}$ criterion.
The selected hyperparameter values are reported in Table~\ref{tab:hyperparameters-list}, and additional tuning results are provided in the Supplementary Information (SI).
The configuration files, hyperparameter search space, and scripts used to run the Optuna study are available in the AutoREC GitHub repository \cite{GitHub}.

\begin{table*}[!hbt]
    \centering
    \resizebox{\textwidth}{!}{
        \begin{threeparttable}
            \caption{
                Hyperparameter search space and the corresponding optimal values for training the RL agent.
            }
            \label{tab:hyperparameters-list}

            \begin{tabular}{l l c c l}
              \hline
              Hyperparameter & Type & Range & Optimal value & Description \\
              \hline
              Batch size ($\times 50$) & Integer & $[1, 10]$ & $2$ & Mini-batch size for network updates \\
              Replay buffer capacity ($\times 1{,}000$) & Integer & $[1, 20]$ & $20$ & Maximum number of stored transitions \\
              Training update frequency & Integer & $[1, 100]$ & $14$ & Episodes between network updates \\
              Target network update frequency ($\times 100$) & Integer & $[1, 50]$ & $5$ & Episodes between target network updates \\
              Initial $\epsilon$ & Float & $[0.5, 1.0]$ & $0.9491$ & Initial exploration probability \\
              $\epsilon$ decay & Float & $[0.8, 0.9999]$ & $0.9643$ & Multiplicative decay rate of $\epsilon$ \\
              Minimum $\epsilon$ & Float & $[0.0, 0.1]$ & $0.0932$ & Lower bound on exploration probability \\
              Prioritized replay $\alpha$ & Float & $[0.5, 1.5]$ & $0.6282$ & Degree of prioritization in replay sampling \\
              Initial $\beta$ & Float & $[0.01, 0.6]$ & $0.1074$ & Initial importance-sampling correction \\
              Final $\beta$\tnote{a} & Float & $[0.7, 1.3]$ & $0.7477$ & Final importance-sampling correction \\
              \hline
            \end{tabular}

            \begin{tablenotes}
                \footnotesize
                \item[a] Although $\beta$ is theoretically bounded by $\beta \leq 1.0$, the present implementation permits values $\beta > 1.0$, which effectively control the rate at which the importance-sampling correction increases during training.
            \end{tablenotes}
        \end{threeparttable}
    }
\end{table*}

\paragraph{Computational cost}
Hyperparameter optimization, agent training, and evaluation were performed using the Digital Research Alliance of Canada's Fir high-performance computing cluster.
The calculations were run on nodes equipped with AMD EPYC\texttrademark~9655 processors.
Each 10,000-episode training run required approximately 20 CPU-hours when executed on a single CPU core.
During hyperparameter optimization, the five trials within each batch were run concurrently on five CPU cores.
Thus, the full 50-trial study required approximately 1,000 CPU-hours in aggregate, corresponding to approximately 200 hours of wall-clock time across the 10 sequential batches.
Once trained, the agent can be applied directly to new spectra without further optimization, requiring less than 33 CPU-seconds per spectrum on the same single-core hardware configuration.
For this reference-agent workflow, much of the computational cost is therefore incurred during the upfront hyperparameter optimization and training stages, after which evaluation of additional spectra is comparatively inexpensive.
For applications in which the supplied reference agent is appropriate, the upfront optimization and training cost has already been incurred, whereas a new application requiring retraining or renewed hyperparameter optimization incurs the corresponding model-development cost again.

A possible approach to improve computational efficiency may involve surrogate-assisted screening could be considered to reduce the number of parameter-fitting evaluations in AutoREC.
However, parameter fitting is an integral part of the sequential RL process, providing the fit-quality information used to calculate rewards and determine termination conditions.
Inaccurate surrogate-based screening could therefore reject promising intermediate circuits and affect subsequent circuit-construction steps, potentially degrading the quality of the resulting ECMs.
Furthermore, the computational benefits would need to justify the additional cost of developing and evaluating the surrogate.
Any final candidate obtained through such an approach would still require verification using the full parameter-fitting procedure.

\subsection{Demonstration on synthetic data}
\label{sec:synthetic-data}

We use the synthetic dataset to demonstrate several capabilities of AutoREC, including comparison of agent-training configurations, batch evaluation of a trained agent, post-processing of generated circuits, and inspection of circuit-construction trajectories.

\begin{figure*}[!hbt]
    \centering
    \includegraphics[width=0.9\textwidth]{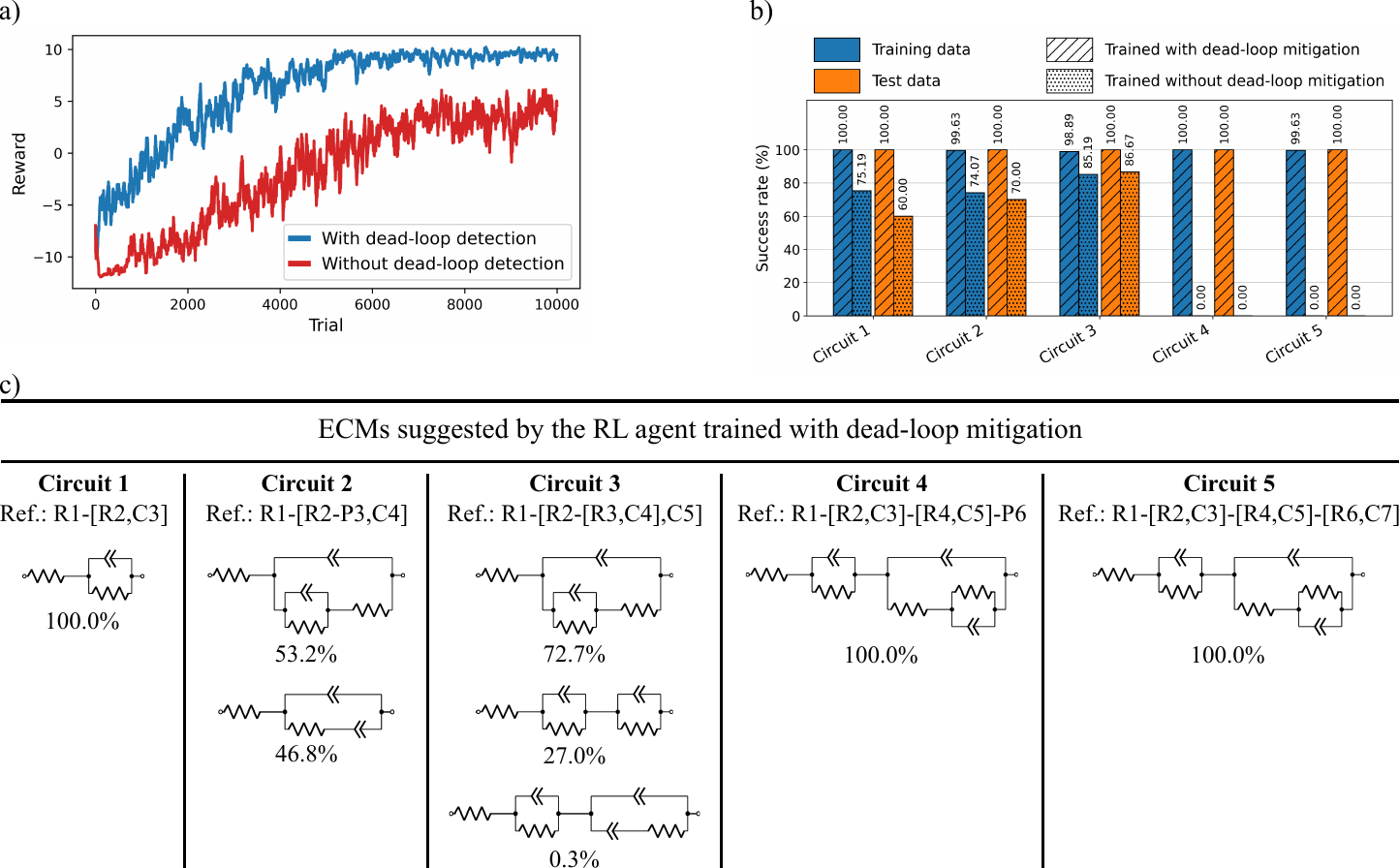}
    \caption{
        Training and evaluation results on the synthetic data, including a comparison of agents trained with and without dead-loop mitigation.
        The agent trained with dead-loop mitigation using the hyperparameter configuration listed in Table~\ref{tab:hyperparameters-list} serves as the reference agent; the comparison agent was trained using the same hyperparameters but without dead-loop mitigation.
        (a) Learning curves for agents trained with (blue) and without (red) the dead-loop mitigation strategy described in Algorithm~\ref{alg:dead-loop}.
        (b) Corresponding success rates on the training and test sets.
        (c) ECMs suggested by the reference agent for each reference circuit topology.
        Within each topology, the percentages indicate how frequently each ECM was suggested among the successful cases across the training and test sets.
    }
    \label{fig:4-combined-training-results}
\end{figure*}

The reference agent used in the subsequent demonstrations was trained according to the procedure described in Sec.~\ref{sec:agent-training} with the dead-loop mitigation strategy described in Algorithm~\ref{alg:dead-loop} enabled.
To assess the effect of this strategy, we compared training with and without dead-loop mitigation using the hyperparameter set listed in Table~\ref{tab:hyperparameters-list}.
We further repeated this comparison across multiple random seeds to assess the consistency of the observed behavior, with the detailed results provided in the SI.
The comparison uses the learning curve and success rate as evaluation metrics.
The learning curve reports the episodic return, smoothed using a 50-episode moving average, while the success rate is defined in Sec.~\ref{sec:agent-training}.
For the synthetic spectra considered here, the data-dependent $\chi^2_{\mathrm{thresh}}$ used to determine success does not exceed $6.23\times10^{-5}$.

As shown in Fig.~\ref{fig:4-combined-training-results}a, enabling dead-loop mitigation resulted in faster reward progression and higher final reward.
The strategy also produced higher success rates across the reference ECM families, as shown in Fig.~\ref{fig:4-combined-training-results}b.
Without mitigation, repeated invalid actions leave the circuit unchanged and can introduce highly redundant transitions into the replay buffer, reducing the diversity of experiences available for training.
By temporarily restricting action selection after a dead loop is detected, the mitigation strategy encourages exploration of valid circuit modifications while retaining the invalid-action experience needed to learn which actions should be avoided.
These results provide empirical support for the dead-loop mitigation strategy implemented in AutoREC as a means of reducing redundant transitions and improving exploration during agent training.
Based on these results, dead-loop mitigation was enabled when training the reference agent evaluated below.

After training, the agent parameters were fixed, and the resulting policy was applied in batch to both the synthetic training spectra and the held-out test spectra without retraining for individual measurements.
AutoREC records the evaluation results in a structured table, including whether a satisfactory ECM was identified, the selected candidate circuit, its fitted parameters and fit-quality metrics, the number of actions performed, and the complete action history.
The generated circuits were subsequently simplified using the post-processing utilities described in Sec.~\ref{sec:postprocessing-analysis} by combining electrically reducible components connected in series or parallel.
The resulting simplified ECMs are summarized in Fig.~\ref{fig:4-combined-training-results}c, with percentages indicating the frequencies of the suggested topologies among successful evaluations pooled across the training and test sets.
A quantitative summary of the reference agent's performance for each synthetic circuit family is provided in Table~\ref{tab:synthetic-evaluation}, including fit-based success, exact topology recovery, circuit complexity, and median fitting errors for each reference circuit family; the corresponding fitting-error distributions is provided in the SI.

\begin{table*}[!hbt]
    \centering
    \caption{
        Quantitative evaluation of the reference agent trained with dead-loop mitigation on the five synthetic reference circuit families.
        Values are reported as training (test) for each reference family.
        Fit success is defined by the data-dependent $\chi^2_{\mathrm{thresh}}$ criterion.
        Exact topology recovery is the percentage of all evaluated spectra whose simplified ECM matches the reference topology, treating capacitors and constant-phase elements as equivalent.
        Circuit complexity is measured by the number of ECM parameters, with generated complexity reported as the median across all valid simplified ECMs.
        Fitting error is summarized by the median $\log_{10}(\chi^2)$ across all evaluations with finite fitting errors.
        These results characterize the specific reference agent evaluated here and should not be interpreted as representative of the performance of other agents developed using AutoREC.
    }
    \label{tab:synthetic-evaluation}

    \renewcommand{\arraystretch}{1.2}
    \setlength{\tabcolsep}{12pt}

    \resizebox{\textwidth}{!}{
    \begin{tabular}{lccccc}
        \toprule
        \textbf{Metric}
        & \textbf{Circuit 1}
        & \textbf{Circuit 2}
        & \textbf{Circuit 3}
        & \textbf{Circuit 4}
        & \textbf{Circuit 5} \\
        \midrule

        Fit-based success (\%)
        & $100.00$ ($100.00$)
        & $ 99.63$ ($100.00$)
        & $ 98.89$ ($100.00$)
        & $100.00$ ($100.00$)
        & $ 99.63$ ($100.00$) \\

        Exact topology recovery (\%)
        & $100.00$ ($100.00$)
        & $ 47.41$ ($ 40.00$)
        & $ 71.85$ ($ 73.33$)
        & $  0.00$ ($  0.00$)
        & $  0.00$ ($  0.00$) \\

        \midrule

        Reference complexity ($k_{\mathrm{ref}}$)
        & $3$ & $5$ & $5$ & $7$ & $7$ \\

        Generated complexity ($k$)
        & $ 4$ ($ 4$)
        & $ 7$ ($ 7$)
        & $ 7$ ($ 7$)
        & $10$ ($10$)
        & $10$ ($10$) \\

        \midrule

        Median $\log_{10}(\chi^2)$
        & $-5.70$ ($-5.71$)
        & $-5.44$ ($-5.42$)
        & $-5.71$ ($-5.71$)
        & $-5.67$ ($-5.67$)
        & $-5.72$ ($-5.71$) \\

        \bottomrule
    \end{tabular}
    }
\end{table*}

Inspection of the simplified ECMs in Fig.~\ref{fig:4-combined-training-results}c reveal consistent circuit-generation patterns within each reference ECM family.
For Circuits~1, 4, and 5, all successful evaluations produce the same simplified topology, whereas Circuits~2 and 3 yield a small number of closely related topologies.
This consistency suggests that the trained policy develops systematic preferences for a small set of recurring circuit topologies rather than producing a broad set of unrelated structures.
However, consistency in the generated structures characterizes the agent's behavior within the training distribution and does not necessarily imply exact recovery of the corresponding reference circuit or generalization to spectra outside this distribution.

Although the generated ECMs reproduced the target impedance spectra with high fitting accuracy, some differed from their corresponding reference circuits.
Several of these differences reflect the inherent non-uniqueness of ECM identification, whereby distinct circuit structures can produce nearly indistinguishable impedance responses \cite{berthier_distinguishability_2001,buteau_explicit_2018,george_robust_2026,orazem_can_2026}.
This ambiguity is intrinsic to impedance-based ECM identification rather than a consequence of the AutoREC formulation.
For example, the two dominant topologies generated for Circuit~3 in Fig.~\ref{fig:4-combined-training-results}c are structurally different but can produce the same impedance response, differing only in whether one RC subcircuit is arranged in series or nested within the other.
The generated ECMs also contained more free parameters than their corresponding reference circuits, partly reflecting the use of CPEs in place of ideal capacitors.
Other generated circuits contained additional elements whose fitted contributions were negligible over the frequency range considered, indicating that the complexity penalty was not always sufficiently effective in promoting the most parsimonious solution.
Because AutoREC exposes the generated topology together with its fitted parameters and fit-quality metrics, users can identify such elements during analysis and potentially address them through contribution-based pruning, additional post-processing rules, or alternative reward formulations.

More generally, the reference agent is designed to identify circuit structures that reproduce the observed impedance response under the specified fitting and complexity objectives.
The resulting ECMs therefore provide candidate representations of the EIS response, while their connection to specific electrochemical mechanisms requires subsequent physical interpretation.
AutoREC automates the search, fitting, and evaluation of these candidate models, while the assignment of physical meaning to the identified circuit elements or processes remains a domain-specific interpretation step.

To provide an external point of reference, we also evaluated AutoEIS on the synthetic data.
Because AutoEIS uses a stochastic GEP-based search and may return multiple candidate circuits after filtering and Bayesian inference, its outputs are not directly equivalent to the single candidate generated by the reference agent; detailed comparisons are therefore provided in the SI.
A notable difference is the computational profile of the two approaches.
Both methods were evaluated using the same hardware and software environment, with computations restricted to a single CPU core.
Once trained, the AutoREC reference agent required less than $33$~CPU-seconds per spectrum for the complete circuit-generation process, including iterative topology modifications and parameter fitting, whereas running the complete AutoEIS workflow, including circuit generation, filtering, Bayesian inference, and model evaluation, required approximately $4$~CPU-hours per spectrum on average under the conditions considered here.
AutoREC therefore incurs substantial computation during the upfront training stage and hyperparameter optimization, as detailed in Sec.~\ref{sec:agent-training}, whereas AutoEIS performs a new stochastic search and analysis for each spectrum.
However, these computational times are specific to the tested configurations and do not imply equivalent fitting quality or establish a general performance ranking between the two approaches.

\begin{figure}[!hbt]
    \centering
    \includegraphics[width=0.45\textwidth]{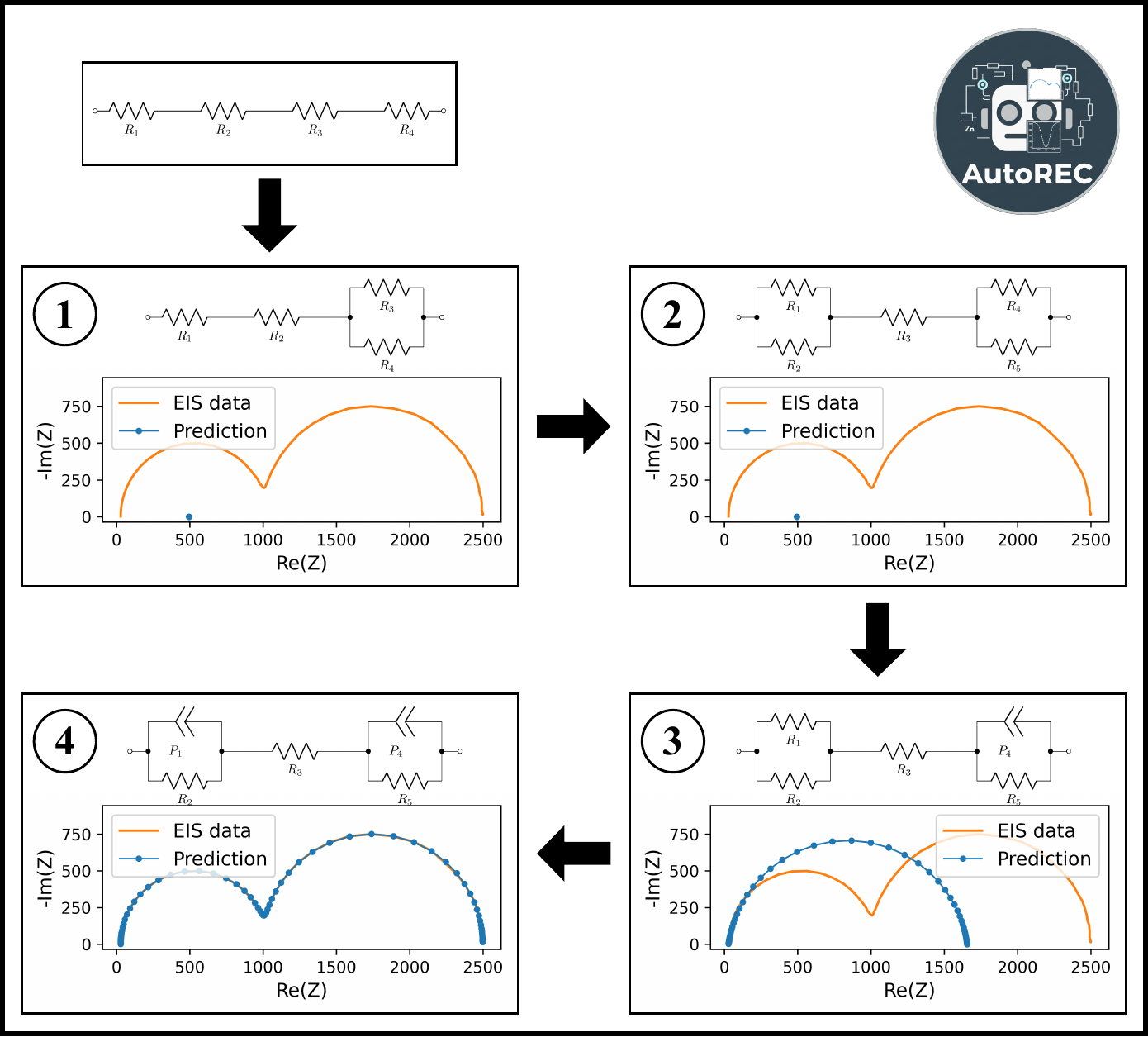}
    \caption{
	An example of the evolution of the ECM and the corresponding Nyquist plots over an episode.
    }
    \label{fig:4-circuit-evolution}
\end{figure}

Beyond the final candidate ECMs, AutoREC retains the complete action history for each evaluation episode, providing access to information such as the intermediate circuit states, rewards, fitted parameters, and fit-quality metrics.
The platform also provides utilities for extracting and visualizing these records, allowing users to inspect how a trained agent incrementally constructs and refines candidate ECMs.
Figure~\ref{fig:4-circuit-evolution} illustrates this capability for a representative spectrum from the synthetic test set by presenting the sequence of intermediate circuit topologies alongside their corresponding fitted Nyquist responses.
Additional examples are provided in the SI.

In this example, the episode begins with an initial circuit composed of series-connected resistors.
During the first few actions, the agent establishes the overall circuit topology by adding connections and forming parallel branches.
It then replaces selected resistors with CPEs, producing Randles-type elements whose fitted responses progressively capture the semicircular features of the target Nyquist spectrum.
By the end of the episode, the resulting ECM closely reproduces the test spectrum.

While the precise mutation sequence varies among spectra, a recurring pattern emerges across the synthetic trajectories examined here: the agent consistently introduces a parallel branch into the initial configuration of series-connected resistor as its first circuit modification.
This physically intuitive heuristic is essential to capture the distinct time constants of electrochemical systems \cite{lasia_2014} and allows the agent to establish the dominant impedance features before refining the model.
More broadly, the trajectory-visualization utilities enable users to identify recurring construction strategies, determine which actions coincide with improvements in reward, and diagnose inefficient or otherwise undesirable circuit-generation sequences.

\subsection{Evaluation on experimental data}
\label{sec:experimental-data}

In addition to synthetic data, AutoREC supports the application and evaluation of trained agents on experimental EIS measurements.
The synthetic-data study provides a controlled setting for developing and evaluating the reference agent, while the practical relevance of the trained policy ultimately depends on how it behaves when applied to real measurements.
Furthermore, experimental spectra commonly contain measurement noise, experimental variability, and impedance features that may not be well represented in the synthetic training data, making them a more stringent test of the learned policy.
Applying the reference agent to these data therefore demonstrates the experimental-data workflow supported by AutoREC while assessing the current agent under more realistic conditions.
Accordingly, this evaluation is intended to serve as a practical demonstration of the software and a means of examining the behavior and limitations of the reference agent, rather than to suggest that autonomous identification of ECMs from experimental data has been fully resolved.

\begin{figure}[!hbt]
    \centering
    \includegraphics[width=0.45\textwidth]{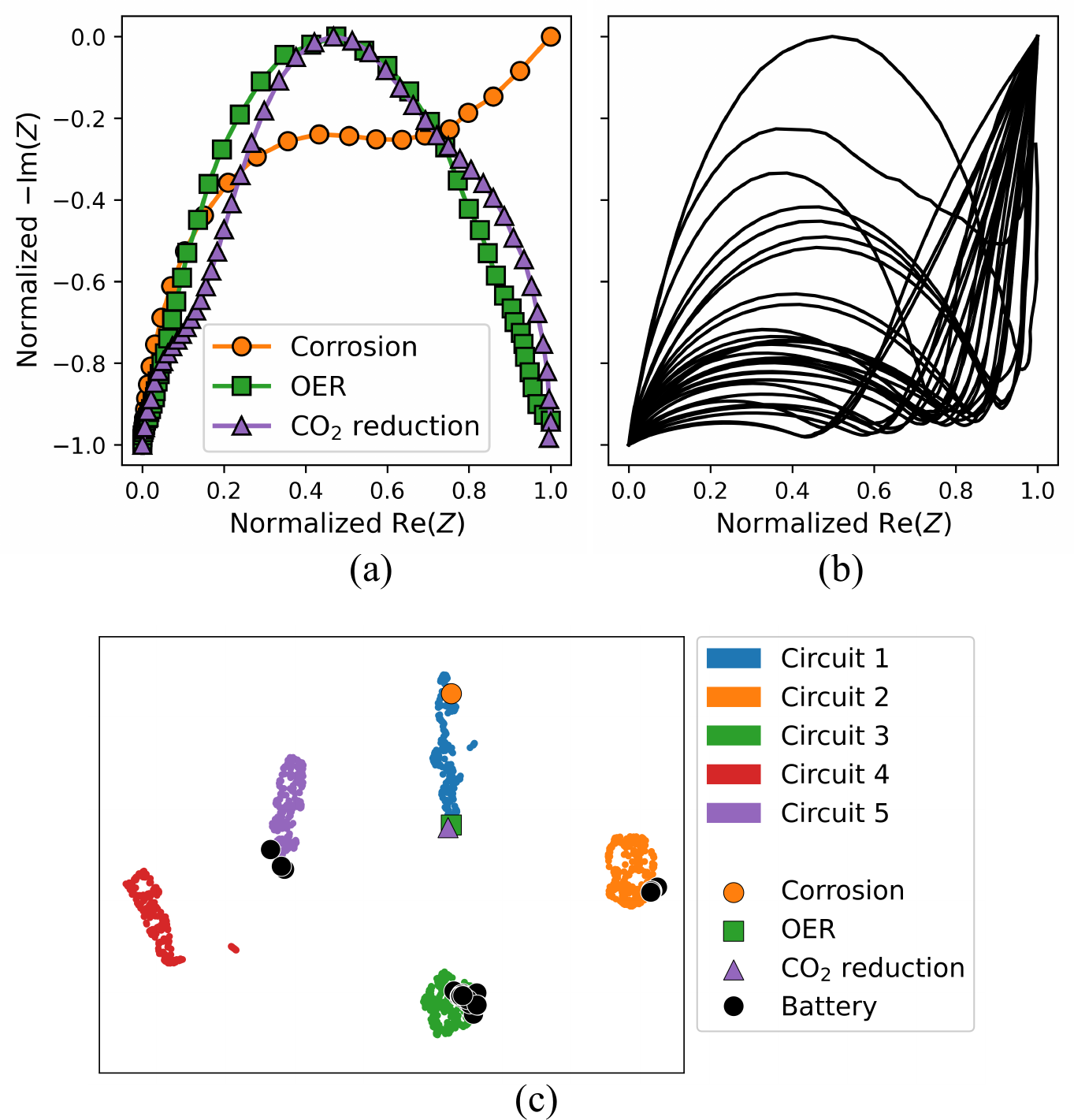}
    \caption{
        Illustration of the experimental EIS datasets used to evaluate the reference agent.
        (a) Normalized Nyquist plots of impedance spectra from corrosion, oxygen evolution reaction (OER), and CO$_2$ reduction systems.
        (b) Normalized Nyquist plots of the battery impedance spectra considered in this study.
        (c) UMAP projection comparing these experimental data with the training set.
    }
    \label{fig:4-experimental-data}
\end{figure}

The experimental dataset includes one representative spectrum each from corrosion, oxygen evolution reaction (OER), and CO$_2$ reduction systems \cite{zhang_editors_2023}, as well as $29$ battery half-cell spectra.
These examples span a range of impedance characteristics and electrochemical applications.
Figure~\ref{fig:4-experimental-data} panels a and b present their normalized Nyquist plots, which allow direct comparison with the synthetic spectra (see Fig.~\ref{fig:4-dataset}).
Panel c then shows a UMAP projection comparing the experimental spectra with the synthetic training data.
The projection provides a qualitative view of how the experimental responses relate to the spectral patterns represented during training.

\begin{figure*}[!hbt]
    \centering
    \includegraphics[width=0.9\textwidth]{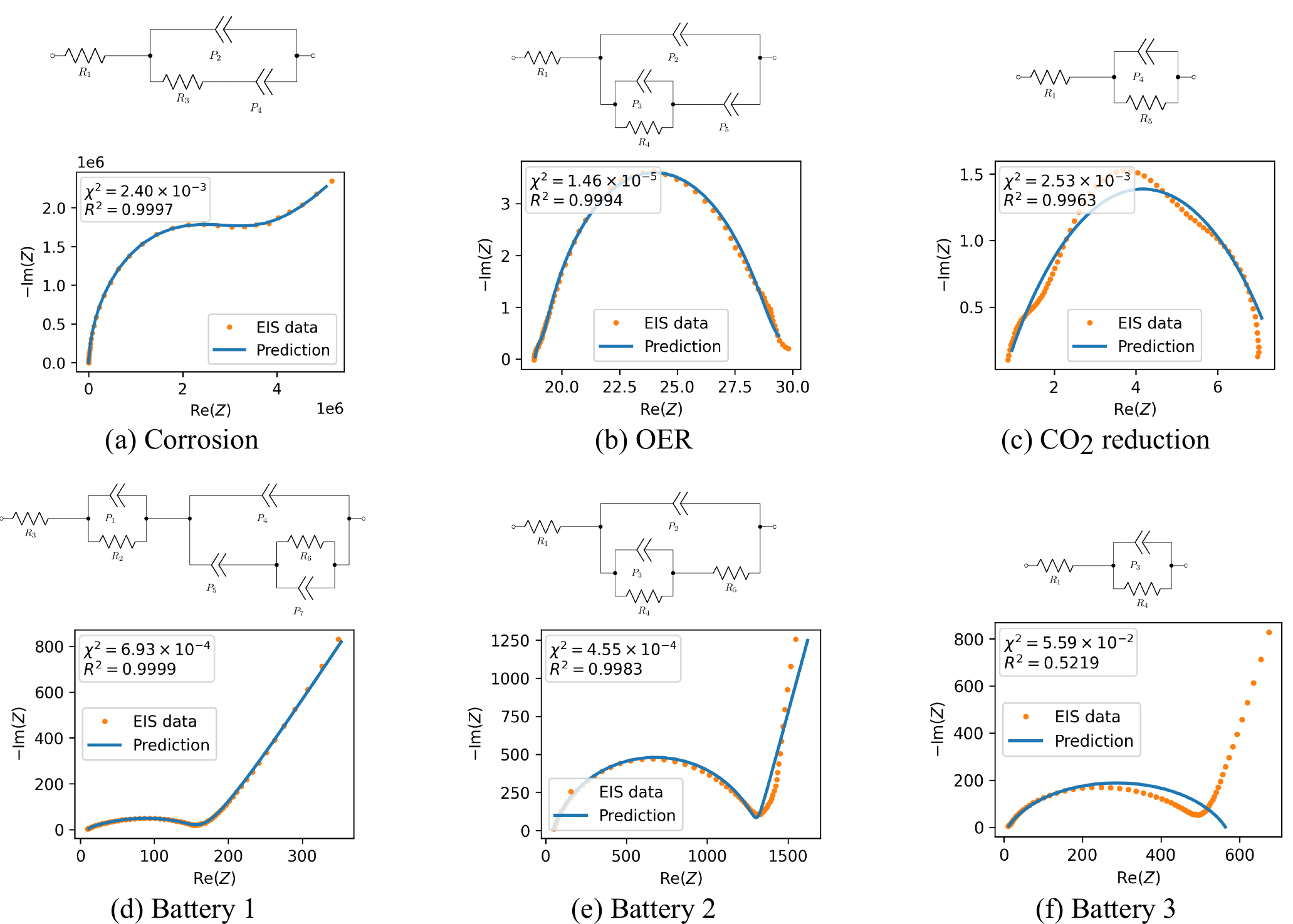}
    \caption{
        ECMs suggested by the reference agent for representative experimental EIS spectra, together with Nyquist plots comparing the experimental spectra (orange points) and the fitted impedance responses of the corresponding suggested ECMs (blue curves).
        The corresponding $\chi^2$ and $R^2$ values are reported in each panel.
        Results are shown for (a) corrosion, (b) oxygen evolution reaction (OER), (c) CO$_2$ reduction, and (d)--(f) battery half-cell systems.
        For the battery data, panels (d) and (e) represent cases in which the suggested ECMs provide high-quality fits to the experimental spectra, whereas panel (f) represents a case in which the suggested ECM is insufficient to reproduce the low-frequency EIS response.
    }
    \label{fig:4-experimental-results}
\end{figure*}

Using this workflow, the reference agent was applied to each experimental spectrum.
AutoREC retains the resulting candidate ECMs, fitted impedance responses, model parameters, and fit-quality metrics for subsequent inspection.
Figure~\ref{fig:4-experimental-results} presents the results for the corrosion, OER, and CO$_2$ reduction spectra together with three representative battery spectra, which shows the suggested ECMs and compares the experimental Nyquist spectra with their corresponding fitted responses.
The corresponding $\chi^2$ and $R^2$ values are also reported in each panel, with the frequency-dependent absolute residuals of the real and imaginary impedance components are provided in the SI.
Results for all $29$ battery spectra are also provided in the SI.

For the corrosion, OER, and CO$_2$ reduction systems, we compare the ECMs suggested by the reference agent with the expert-constructed ECMs and AutoEIS results reported previously in \citet{zhang_editors_2023}.
These comparisons provide an external point of reference for interpreting the reference agent and illustrate how AutoREC supports the development and evaluation of automated ECM-construction agents.
They show what this reference agent can achieve and where its limitations lie, rather than establishing a ranking against AutoEIS.
Beyond topology and fit quality, Bayesian inference is used to examine how well the parameters of the agent-suggested ECMs are constrained by the experimental data, with posterior analyses provided in the SI.

For the corrosion spectrum, the reference agent suggests the same circuit topology as the ECM prioritized by AutoEIS, and the fitted response captures both the main relaxation feature and the low-frequency diffusion-like response.
For OER, the suggested ECM closely reproduces both the dominant semicircular response and the smaller high-frequency feature.
The resulting circuit contains additional structure beyond the single-Randle model prioritized by AutoEIS and, in this respect, more closely resembles the expert-constructed model, which includes a second Randle element.
For CO$_2$ reduction, by contrast, the reference agent suggests an ECM containing only one Randle element, which is simpler than both the expert-constructed model and the simplified model evaluated in the previous AutoEIS study.
The fitted response exhibits only a single semicircular pattern, with visible deviations from the experimental spectrum.
This result illustrates that the current reference agent can capture the dominant behavior while failing to resolve finer features in some experimental spectra.

The battery results extend this evaluation to a larger set of experimental spectra.
AutoEIS was also applied to the battery spectra as an external point of reference, with detailed results provided in the SI.
Although the battery spectra broadly share two dominant features---a semicircular relaxation and a lower-frequency response that often resembles a diffusion-like tail---the reference agent shows different behaviors across the dataset.
For $19$ of the $29$ spectra, the AutoREC reference agent suggests ECMs that capture both features with $\chi^2$ on the order of $10^{-3}$ or lower.
Even among these better-fitting cases, the suggested circuit topology varies.
For the remaining spectra, the results are more limited.
In some cases, the reference agent suggests an ECM that represents the response primarily as a single semicircle, whereas in others it fails to progress beyond its initial circuit state.
Figure~\ref{fig:4-experimental-results}f illustrates the single-semicircle case, where the fitted response fails to reproduce the lower-frequency feature.

The experimental results reveal a limitation of the current reference agent that is not apparent from the synthetic-data evaluation alone.
Many experimental spectra share similar broad features in the Nyquist representation, yet the agent suggests good-fitting ECMs for some spectra and more limited results for others.
This behavior is difficult to understand from the impedance spectra alone because the DDQN acts on a learned representation of the impedance input and circuit state rather than directly on the Nyquist plot.
Although we do not examine this latent representation here, the UMAP projection in Fig.~\ref{fig:4-experimental-data}c provides a complementary view of the impedance-derived features supplied to the agent.
In this representation, the synthetic training data form distinct clusters, while many experimental spectra lie near their peripheries.
This suggests that some experimental inputs may occupy regions that are less well covered by the training distribution, which may make it more difficult for the agent to learn how to handle them.
The UMAP projection does not explain why the agent succeeds or fails on an individual spectrum, but it suggests that training-data coverage may contribute to the observed variation in behavior.

A possible mitigation is to improve the coverage of the training distribution.
The training set could be diversified to span a broader range of EIS responses, including cases between the regions represented by the current synthetic data.
Alternatively, when the intended application is known, the dataset could be enriched specifically in the regions relevant to the expected experimental spectra.
In either case, the objective is to provide better training coverage where the agent is expected to operate.
AutoREC can then be used to train new agents on the revised dataset and evaluate whether their behavior improves on the experimental data.

Improved training coverage, however, does not by itself indicate how much confidence should be placed in the ECM suggested for an individual spectrum.
The present results are obtained from a single trained agent, so an apparently good result provides little information about whether independently trained agents would reach similar conclusions.
A future extension could therefore use AutoREC to train and evaluate a committee of agents under different training configurations or other variations.
Consistent predictions across the committee would increase confidence in the result, whereas disagreement could flag spectra for further inspection.

These strategies are not implemented in the present study and are left for future work.
They nevertheless illustrate how AutoREC can support an iterative workflow in which limitations identified during evaluation motivate changes to the training dataset or agent configuration, followed by retraining and reevaluation within the same platform.


\section{Conclusion}
\label{sec:conclusion}

In this work, we introduced AutoREC, an open-source Python library for developing, training, deploying, evaluating, and analyzing RL agents for automated ECM generation from EIS data.
Its central premise is that ECM construction can be learned as a reusable policy, allowing a trained agent to be applied to new measurements without retraining for each individual spectrum.
The reference DDQN case study demonstrates this workflow across synthetic and experimental data while illustrating how AutoREC can expose not only the final suggested ECMs but also the trajectories and intermediate results through which they are constructed.
By separating these capabilities from any individual agent implementation, AutoREC provides a common platform for developing and comparing alternative RL approaches to ECM generation.

More broadly, bringing ECM generation to this scale changes the role that ECM-based EIS analysis can play in electrochemical experimentation.
Rather than remaining a separate analysis performed after data collection, candidate models and fitted parameters can accompany measurements throughout an experimental campaign, providing structured information as new materials and conditions are explored.
In a self-driving laboratory, this creates a path for EIS-derived models to enter the feedback loop that guides what to measure or explore next, moving ECM-based analysis from an endpoint of characterization toward an active component of experimental design and discovery.

Achieving this goal will require continued development of both the agents and the information used to train and evaluate them.
Future work can use AutoREC to investigate broader or application-specific training distributions, alternative reward formulations and agent architectures, uncertainty-aware or ensemble approaches, and adaptive learning strategies that improve agents as new measurements become available.
Because these developments can be implemented and evaluated within the same framework, AutoREC provides a foundation for iteratively improving automated ECM-construction methods rather than treating any particular agent as a final solution.
More generally, the formulation explored here suggests that reinforcement learning may provide a useful framework for scientific problems in which an interpretable model must be assembled through a sequence of structured decisions.
In this sense, AutoREC represents a step toward autonomous scientific workflows that move beyond automated measurement and prediction toward the construction and refinement of models that can support scientific reasoning and discovery.


\section*{Acknowledgements}
This research was funded by the Government of Canada under the National Research Council Canada's Collaborative Science, Technology and Innovation Program (CSTIP) Critical Battery Materials Initiative (CBMI).
The authors also acknowledge the computational resources provided by the Digital Research Alliance of Canada through the Fir cluster.
This work was not conducted by, for, or on behalf of Lila Sciences.

\printcredits

\appendix
\section{Supplementary Information}
The Supplementary Information includes additional hyperparameter tuning results, further examples of circuit topology evolution during episodes for test-set samples, and the ECMs proposed by the demonstrated agent for the battery spectra.

\section*{Data availability}
The code for the AutoREC package is available on GitHub
(\href{https://github.com/AUTODIAL/AutoREC}{https://github.com/AUTODIAL/AutoREC}) \cite{GitHub},
with commit c63e63d19b46747d4cf2e97c76ad9286e500ac8f corresponding to the version used in this study.
The data and results required to reproduce the study are archived at Zenodo
(DOI: \href{https://doi.org/10.5281/zenodo.19987300}{10.5281/zenodo.19987300}).


\bibliographystyle{cas-model2-names}

\bibliography{references.bib, references_zotero.bib}




\end{document}


\title{Supplementary Material for ``AutoREC: A reinforcement learning agent to generate equivalent circuit models for Electrochemical Impedance Spectroscopy''}

\author{Ali Jaberi}
\thanks{These authors contributed equally to this work.}
\affiliation{Clean Energy Innovation Research Centre, National Research Council Canada, Mississauga, ON, Canada}
\author{Yonatan Kurniawan}
\thanks{These authors contributed equally to this work.}
\affiliation{Department of Materials Science and Engineering, University of Toronto, Toronto, ON, Canada}
\author{Robert Black}
\affiliation{Clean Energy Innovation Research Centre, National Research Council Canada, Mississauga, ON, Canada}
\author{Shayan Mousavi M.}
\affiliation{Clean Energy Innovation Research Centre, National Research Council Canada, Mississauga, ON, Canada}
\author{Kabir Verma}
\affiliation{Cheriton School of Computer Science, University of Waterloo, Waterloo, ON, Canada}
\author{Zoya Sadighi}
\affiliation{Clean Energy Innovation Research Centre, National Research Council Canada, Mississauga, ON, Canada}
\author{Santiago Miret}
\affiliation{Lila Sciences, San Francisco, CA, USA}
\author{Jason Hattrick-Simpers}
\email{jason.hattrick.simpers@utoronto.ca}
\affiliation{Department of Materials Science and Engineering, University of Toronto, Toronto, ON, Canada}

\date{\today}

\pacs{}

\maketitle 

\tableofcontents

\section{Additional hyperparameter tuning results}
\label{sec:hyperparameter-tuning}

\begin{figure}[!hbt]
    \centering
    \includegraphics[width=0.9\textwidth]{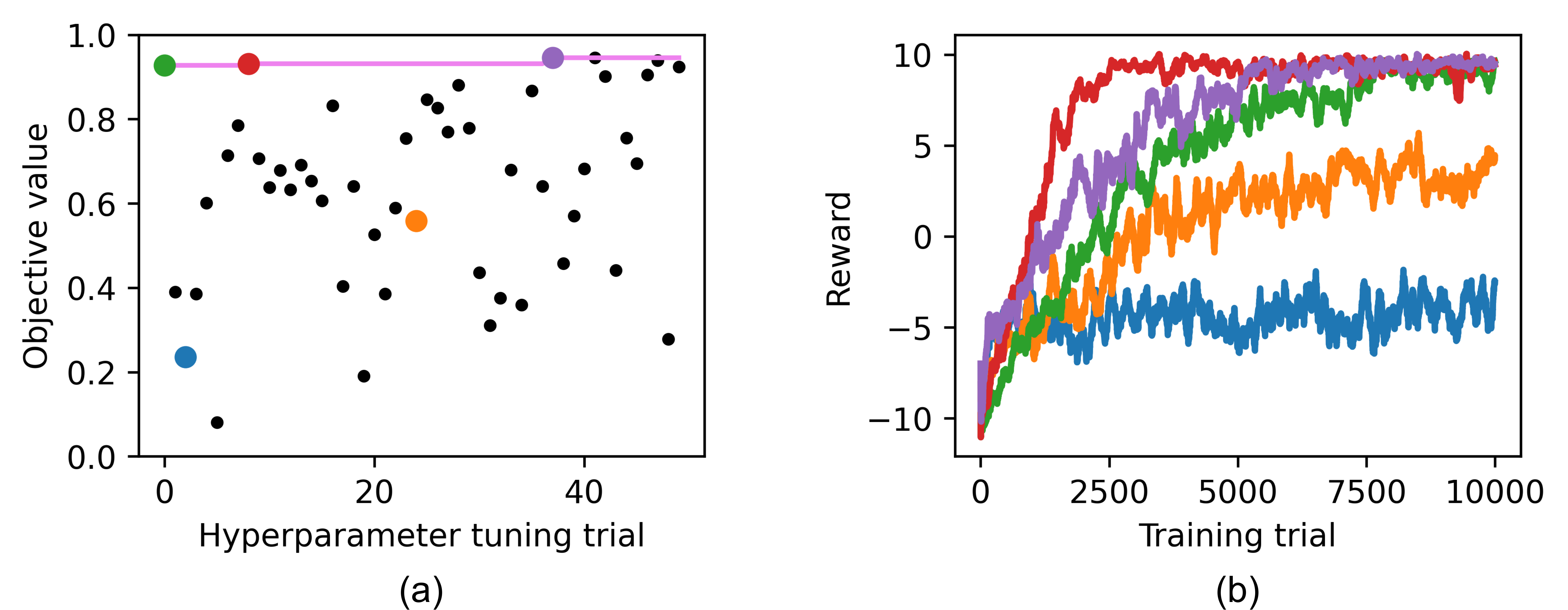}
    \caption{
        (a) Objective values obtained across the hyperparameter-tuning trials, with selected configurations highlighted in color.
        (b) Learning curves of the selected agent configurations, with colors corresponding to the highlighted configurations in panel (a).
        The selected configurations include low-, medium-, and several high-performing configurations.
    }
    \label{fig:hyperparameter-tuning-curves}
\end{figure}

The choice of hyperparameter values can substantially influence agent performance, underscoring the importance of hyperparameter tuning.
To illustrate this effect, we highlight five representative agent configurations obtained during the tuning process, spanning different objective values and learning behaviors.
The hyperparameter values for these configurations are listed in Table~\ref{tab:hyperparameter-values}, while their locations within the hyperparameter search and corresponding learning curves are shown in Fig.~\ref{fig:hyperparameter-tuning-curves}.

\begin{table}[!hbt]
    \centering
    \caption{
        Hyperparameter values for the five selected agent configurations highlighted in Fig.~\ref{fig:hyperparameter-tuning-curves}.
        The column colors correspond to the configurations highlighted in panel (a) and their learning curves shown in panel (b).
    }
    \label{tab:hyperparameter-values}
    \setlength{\tabcolsep}{9pt}
    \begin{tabular}{lrrrrr}
      \toprule
      & \makecell[c]{Conf. 1\\(blue)}
      & \makecell[c]{Conf. 2\\(orange)}
      & \makecell[c]{Conf. 3\\(green)}
      & \makecell[c]{Conf. 4\\(red)}
      & \makecell[c]{\textbf{Selected}\\(purple)} \\
      \midrule
      Batch size ($\times 50$) & 1 & 5 & 3 & 3 & \textbf{2} \\
      Replay buffer capacity ($\times 1{,}000$) & 20 & 14 & 15 & 13 & \textbf{20} \\
      Training update frequency & 84 & 63 & 5 & 3 & \textbf{14} \\
      Target network update frequency ($\times 100$) & 11 & 8 & 10 & 3 & \textbf{5} \\
      Initial $\epsilon$ & 0.5909 & 0.7634 & 1.0000 & 0.9957 & \textbf{0.9491} \\
      $\epsilon$ decay & 0.8367 & 0.8676 & 0.9997 & 0.9992 & \textbf{0.9643} \\
      Minimum $\epsilon$ & 0.0304 & 0.0510 & 0.1000 & 0.9856 & \textbf{0.0932} \\
      Prioritized replay $\alpha$ & 1.0248 & 0.9657 & 0.6000 & 0.5227 & \textbf{0.6282} \\
      Initial $\beta$ & 0.2648 & 0.5972 & 0.4000 & 0.5900 & \textbf{0.1074} \\
      Final $\beta$ & 0.8747 & 1.0400 & 0.7000 & 1.2976 & \textbf{0.7477} \\
      \bottomrule
    \end{tabular}
\end{table}

Inspection of the selected configurations reveals several trends in the hyperparameter combinations associated with different levels of agent performance.
Configurations with larger objective values generally used higher initial $\epsilon$ values and slower $\epsilon$ decay, together with smaller prioritized replay $\alpha$ values and more frequent training updates.
In contrast, batch size, replay-buffer capacity, target-network update frequency, and the $\beta$ schedule do not exhibit a clear trend across the selected configurations.
The minimum $\epsilon$ also varies substantially among configurations with similarly high objective values, indicating that different exploration schedules can lead to strong learning performance.
Overall, these observations suggest that agent performance depends on interactions among multiple hyperparameters rather than on a single narrowly defined set of values.

\section{Effect of dead-loop mitigation across random seeds}
\label{sec:dead-loop}

Dead-loop mitigation algorithm is introduced to reduce repeated invalid transitions during training.
When a dead loop is detected, the available actions are temporarily restricted to valid circuit modifications, allowing the replay buffer to accumulate a more diverse set of transitions instead of repeatedly storing transitions that leave the circuit unchanged.
The main manuscript compares representative agents trained with and without mitigation using the selected hyperparameter configuration in the final column of Table~\ref{tab:hyperparameter-values}.
Here, we repeat this comparison across five random seeds for each condition to examine whether the observed effect is consistent across independent training runs.

\begin{figure}[!hbt]
    \centering
    \includegraphics[width=0.95\textwidth]{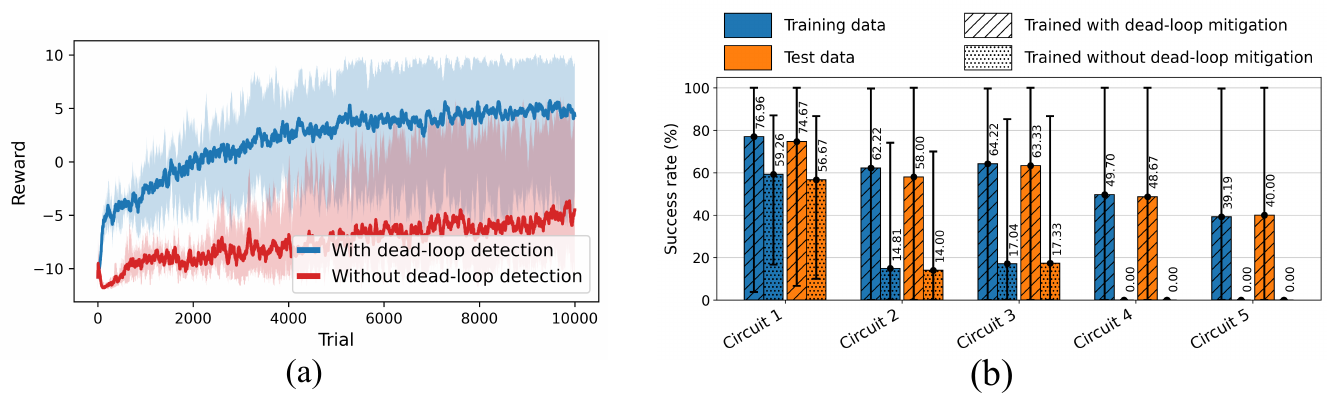}
    \caption{
        Effect of dead-loop mitigation across five random seeds.
        (a) Learning curves for agents trained with (blue) and without (red) dead-loop mitigation.
        The solid lines show the mean episodic return across the five runs, while the shaded regions indicate the corresponding minimum-to-maximum range.
        (b) Success rates on the synthetic training and test sets for each reference circuit family.
        The bars show the mean success rate across the five runs, and the error bars indicate the minimum-to-maximum range.
    }
    \label{fig:deadloop-comparison}
\end{figure}

Figure~\ref{fig:deadloop-comparison}a shows the mean learning curve across the five runs for agents trained with (blue) and without (red) dead-loop mitigation, together with the corresponding minimum-to-maximum envelope.
On average, training with dead-loop mitigation produces faster reward growth and reaches a higher final episodic return.
As expected for stochastic RL training, the performance varies across random seeds and the ranges for the two conditions overlap.
Nevertheless, the mean learning curves favor dead-loop mitigation, indicating that the strategy improves the overall training behavior rather than guaranteeing a better outcome for every individual run.
The success-rate results in Fig.~\ref{fig:deadloop-comparison}b show the same overall trend, with agents trained with dead-loop mitigation achieving higher success rates on average across the synthetic circuit families.

\section{Fitting-error distributions for synthetic data}
\label{sec:synthetic-fitting-errors}

\begin{figure}[!hbt]
    \centering
    \includegraphics[width=\textwidth]{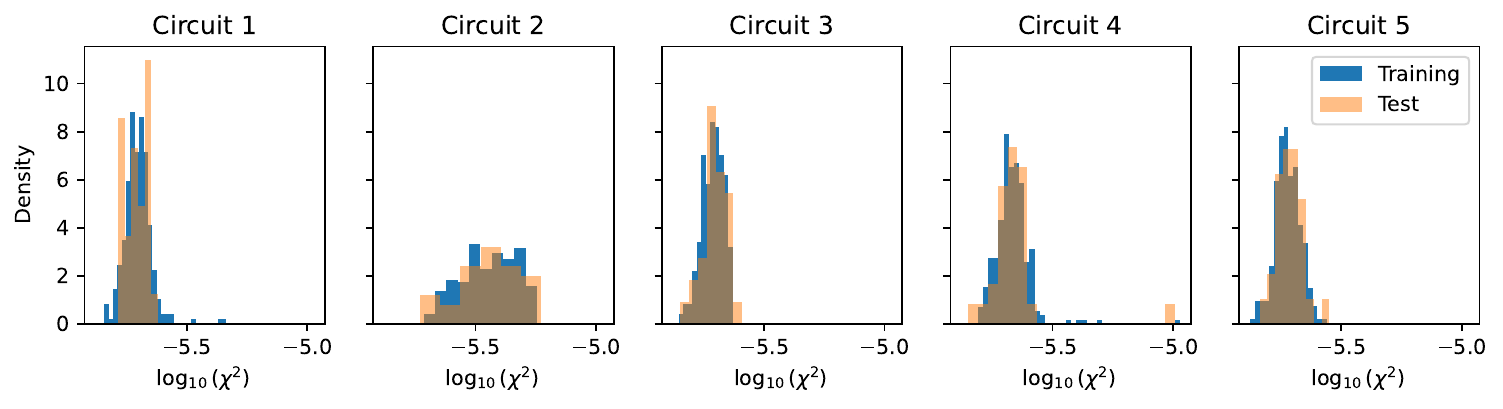}
    \caption{
        Distributions of $\log_{10}(\chi^2)$ for successful evaluations of the reference agent on the five synthetic circuit families.
        Blue and orange histograms represent the training and test sets, respectively, with each distribution normalized to unit area.
        Unsuccessful evaluations are excluded.
    }
    \label{fig:synthetic-fitting-errors}
\end{figure}

Figure~\ref{fig:synthetic-fitting-errors} shows the distributions of $\log_{10}(\chi^2)$ for successful evaluations of the reference agent on the five synthetic circuit families, separately for the training and test sets.
The distributions show substantial overlap between the training and test sets, although their widths and central values vary across the reference circuit families.
Unsuccessful evaluations, which yielded substantially larger fitting errors with $\chi^2$ on the order of $1$, are excluded to better visualize the distributions of successful evaluations.
The corresponding fit-based success rates and median fitting errors are summarized in the main manuscript.

\section{Examples of circuit topology evolution}
\label{sec:circuit-evolution}

\begin{figure}[!hbt]
    \centering
    \includegraphics[scale=0.5]{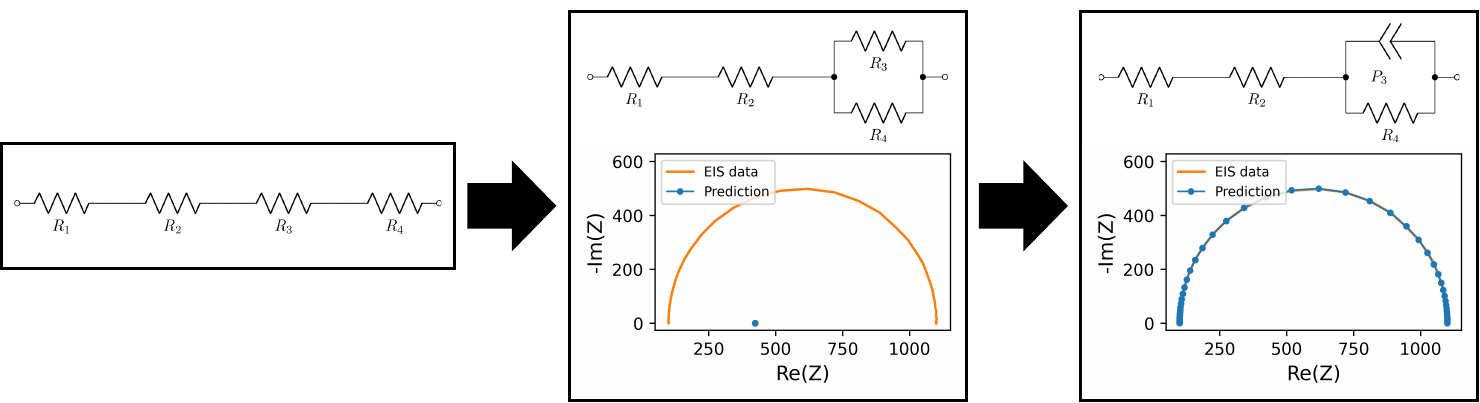}
    \caption{
	Circuit evolution proposed by the trained RL agent during an episode for a representative EIS sample with a single Randles-type element (reference circuit 1).
    }
    \label{fig:circuit-evo-a}
\end{figure}

\begin{figure}[!hbt]
    \centering
    \includegraphics[scale=0.5]{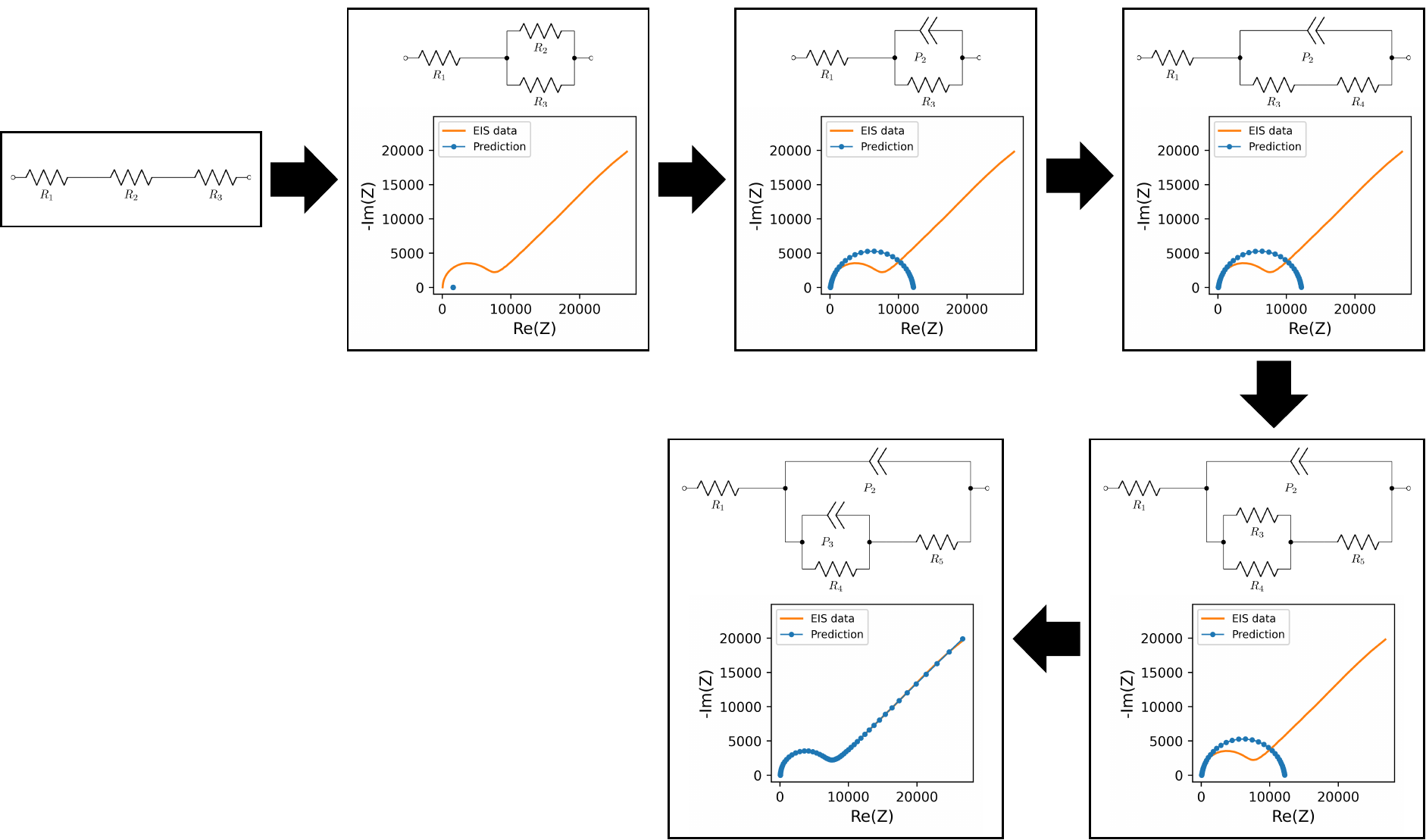}
    \caption{
	Circuit evolution proposed by the trained RL agent during an episode for a representative EIS sample with a single Randles-type element and a Warburg diffusion line (reference circuit 2).
    }
    \label{fig:circuit-evo-b}
\end{figure}

\begin{figure}[!hbt]
    \centering
    \includegraphics[scale=0.5]{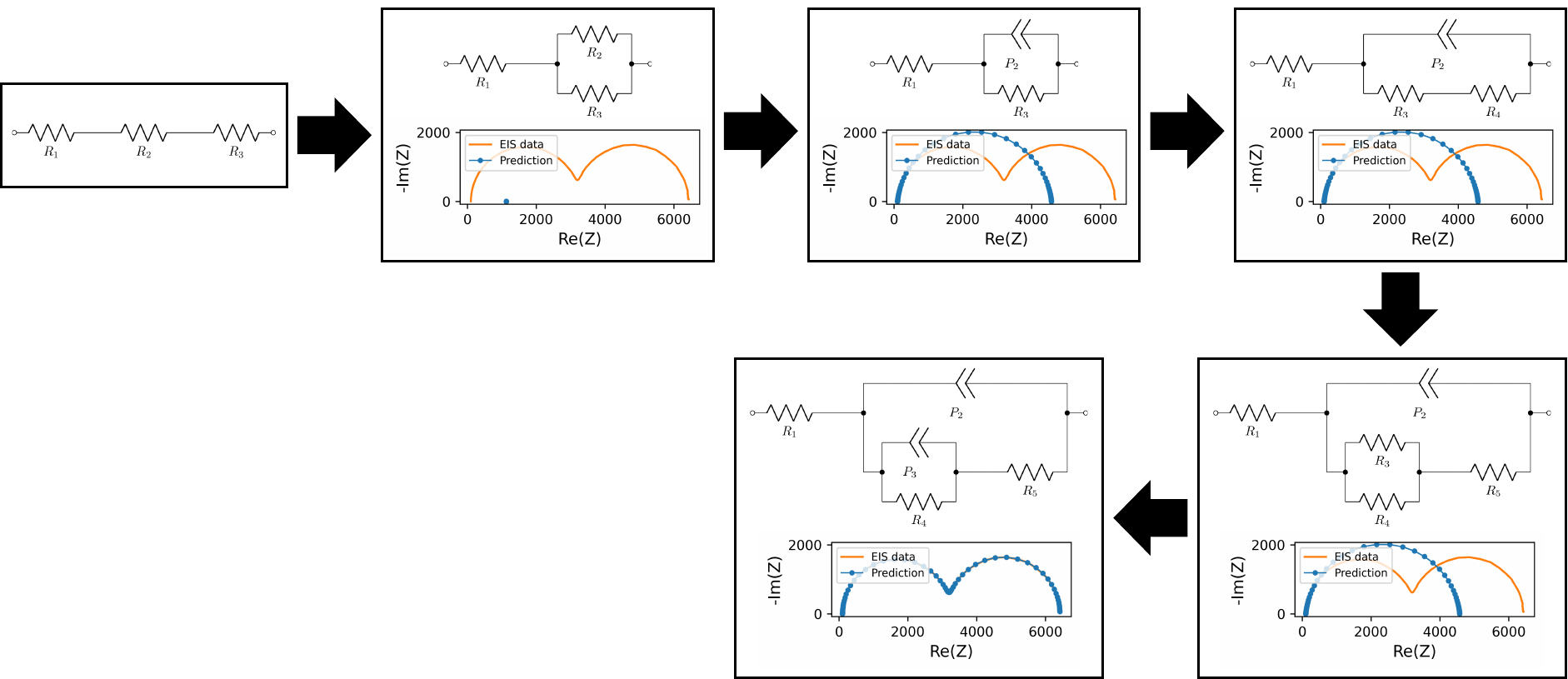}
    \caption{
	Circuit evolution proposed by the trained RL agent during an episode for a representative EIS sample with two Randles-type elements (reference circuit 3).
    }
    \label{fig:circuit-evo-c}
\end{figure}

\begin{figure}[!hbt]
    \centering
    \includegraphics[scale=0.5]{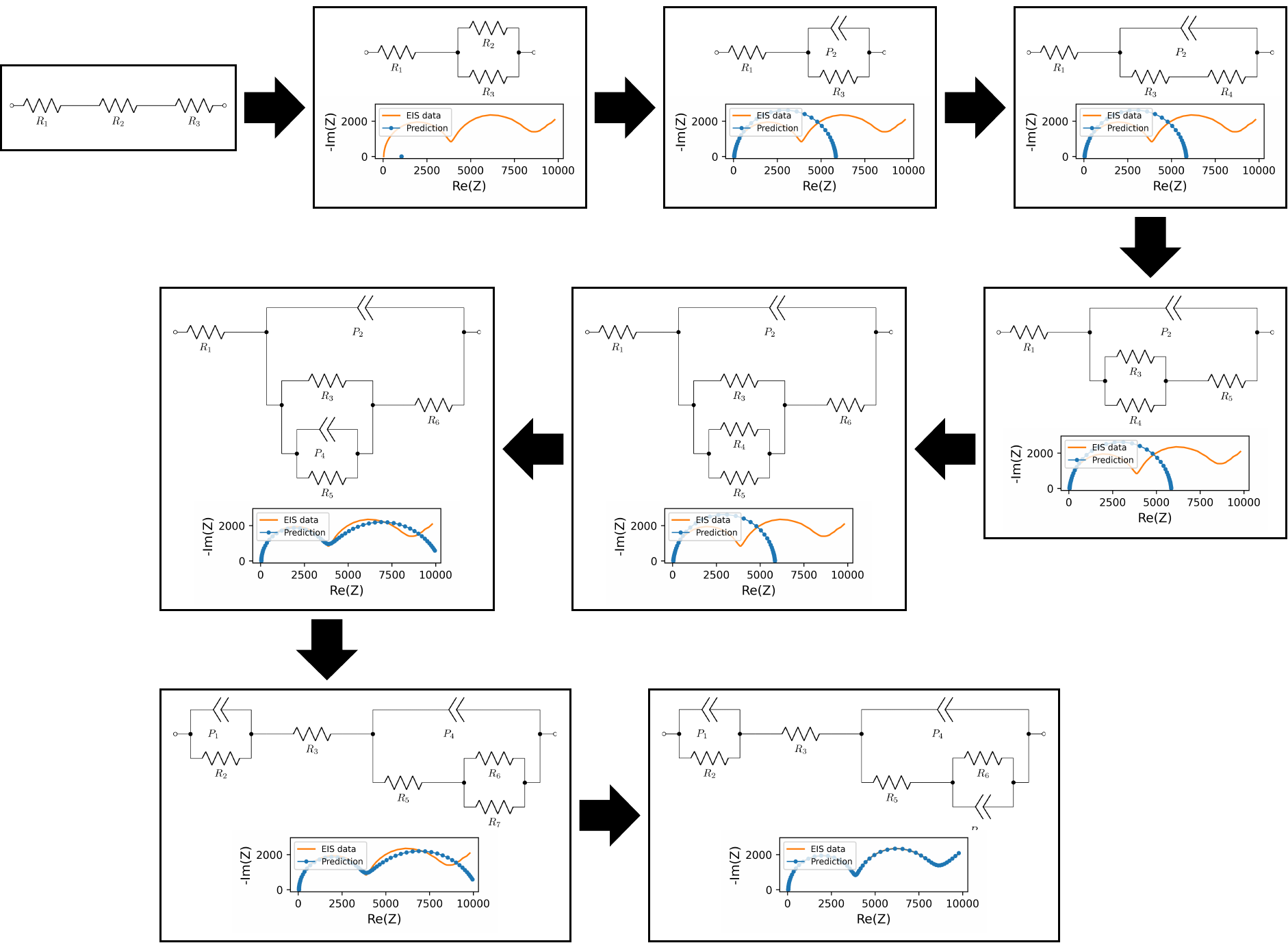}
    \caption{
	Circuit evolution proposed by the trained RL agent during an episode for a representative EIS sample with two Randles-type elements and a Warburg diffusion line (reference circuit 4).
    }
    \label{fig:circuit-evo-d}
\end{figure}

\begin{figure}[!hbt]
    \centering
    \includegraphics[scale=0.5]{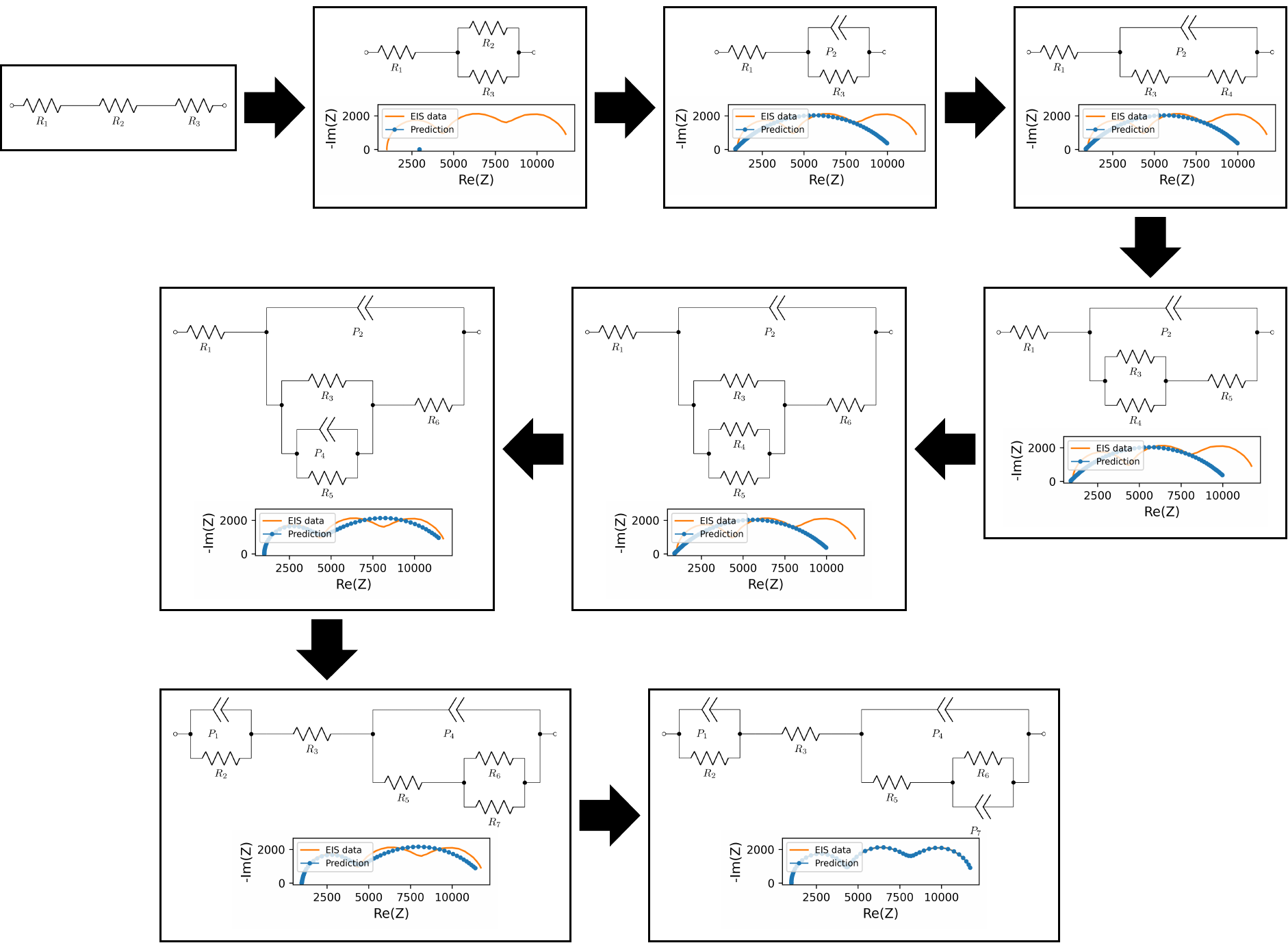}
    \caption{
	Circuit evolution proposed by the trained RL agent during an episode for a representative EIS sample with three Randles-type elements (reference circuit 5).
    }
    \label{fig:circuit-evo-e}
\end{figure}

In this section, we present representative examples of circuit topology evolution during an episode, illustrating how the RL agent modifies the state at each step to arrive at its final suggested ECM.
Figures~\ref{fig:circuit-evo-a}--\ref{fig:circuit-evo-e} show representative evolution sequences generated by the reference agent for one test-set sample corresponding to each of the five reference topologies.

These figures show that the agent suggests targeted modifications that efficiently capture key features of the EIS data and improve the fit.
Interestingly, across all synthetic spectra, the agent always introduces a parallel connection as the initial modification to the circuit.
This heuristic provides a necessary foundation for subsequently constructing a Randles-type subcircuit, which is an important feature present across the synthetic spectra.
Beyond this initial step, the modifications vary across spectra, yet all examples ultimately lead to well-fitting ECMs.

\clearpage

\clearpage

\section{Frequency-dependent residuals for experimental spectra}
\label{sec:experimental-residuals}

\begin{figure}[!hbt]
    \centering
    \includegraphics[width=\textwidth]{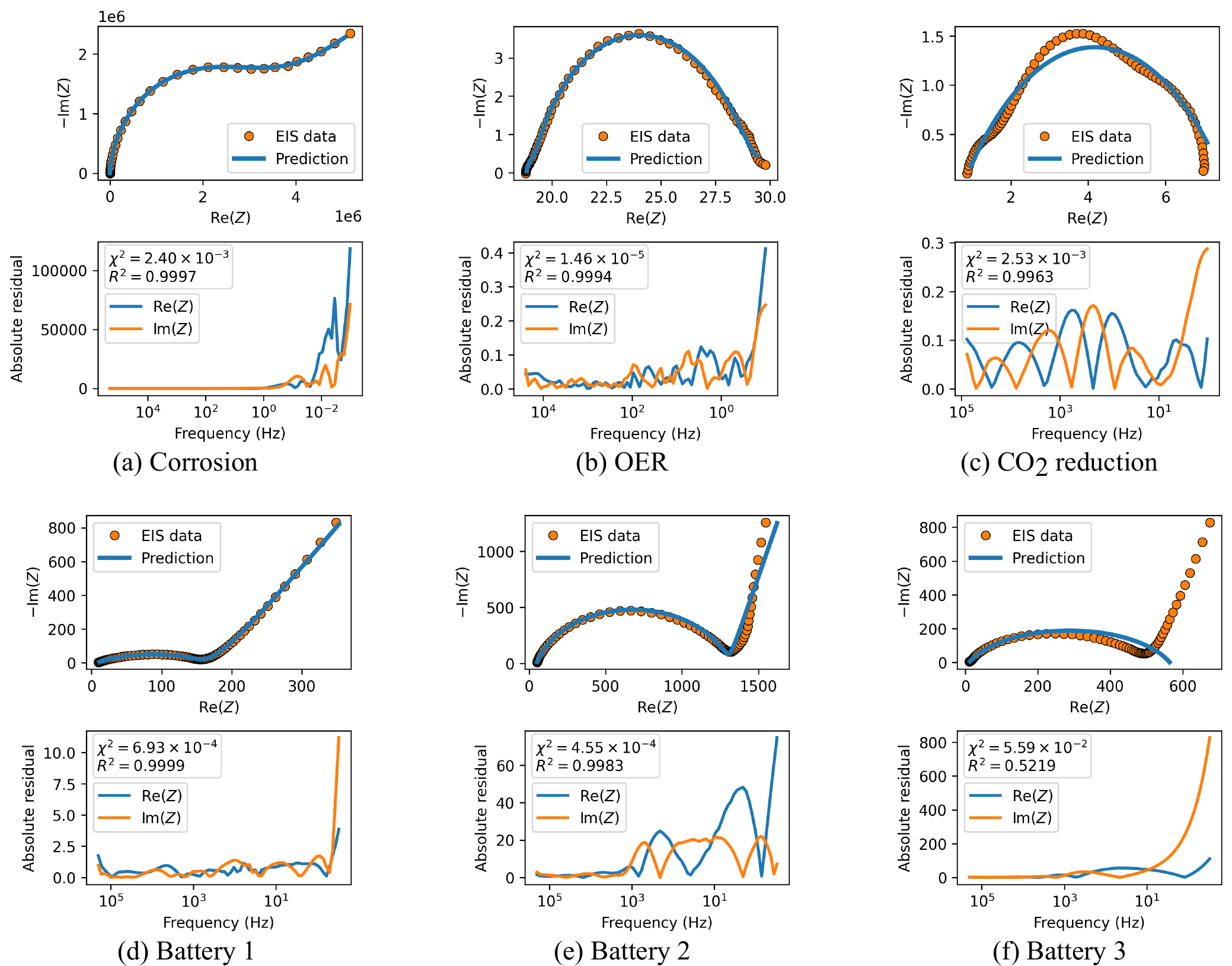}
    \caption{
        Frequency-dependent fitting residuals for the six representative experimental EIS spectra:
        (a) corrosion, (b) OER, (c) CO$_2$ reduction, and (d)--(f) three battery examples.
        The upper panels compare the measured (orange markers) and predicted (blue lines) Nyquist spectra.
        The lower panels show the absolute residuals of the real (blue) and imaginary (orange) impedance components as functions of frequency.
        The frequency axis decreases from left to right to facilitate comparison with the Nyquist plots.
        The corresponding $\chi^2$ and $R^2$ values are also reported.
    }
    \label{fig:experimental-residuals}
\end{figure}

Figure~\ref{fig:experimental-residuals} presents the frequency-dependent fitting residuals for the six representative experimental examples discussed in the main manuscript.
The absolute residuals are calculated as $|\mathrm{Re}(Z_{\mathrm{fit}}-Z_{\mathrm{exp}})|$ and $|\mathrm{Im}(Z_{\mathrm{fit}}-Z_{\mathrm{exp}})|$ for the real and imaginary impedance components, respectively.
The frequency axis is arranged in descending order to facilitate comparison with the Nyquist plots, where higher-frequency data generally appear toward the left for these spectra.

The absolute residuals generally increase toward lower frequencies, indicating more pronounced fitting discrepancies in these regions.
This behavior may reflect difficulties in reproducing low-frequency impedance features, although it may also partly result from the larger impedance magnitudes at lower frequencies.
Because the experimental systems have substantially different impedance scales, the absolute residual magnitudes should not be compared directly across the different examples.

\clearpage
\section{Agent-proposed ECMs for experimental battery spectra}
\label{sec:ecm-battery}

\begin{figure}[!hbt]
    \centering
    \includegraphics[scale=0.33]{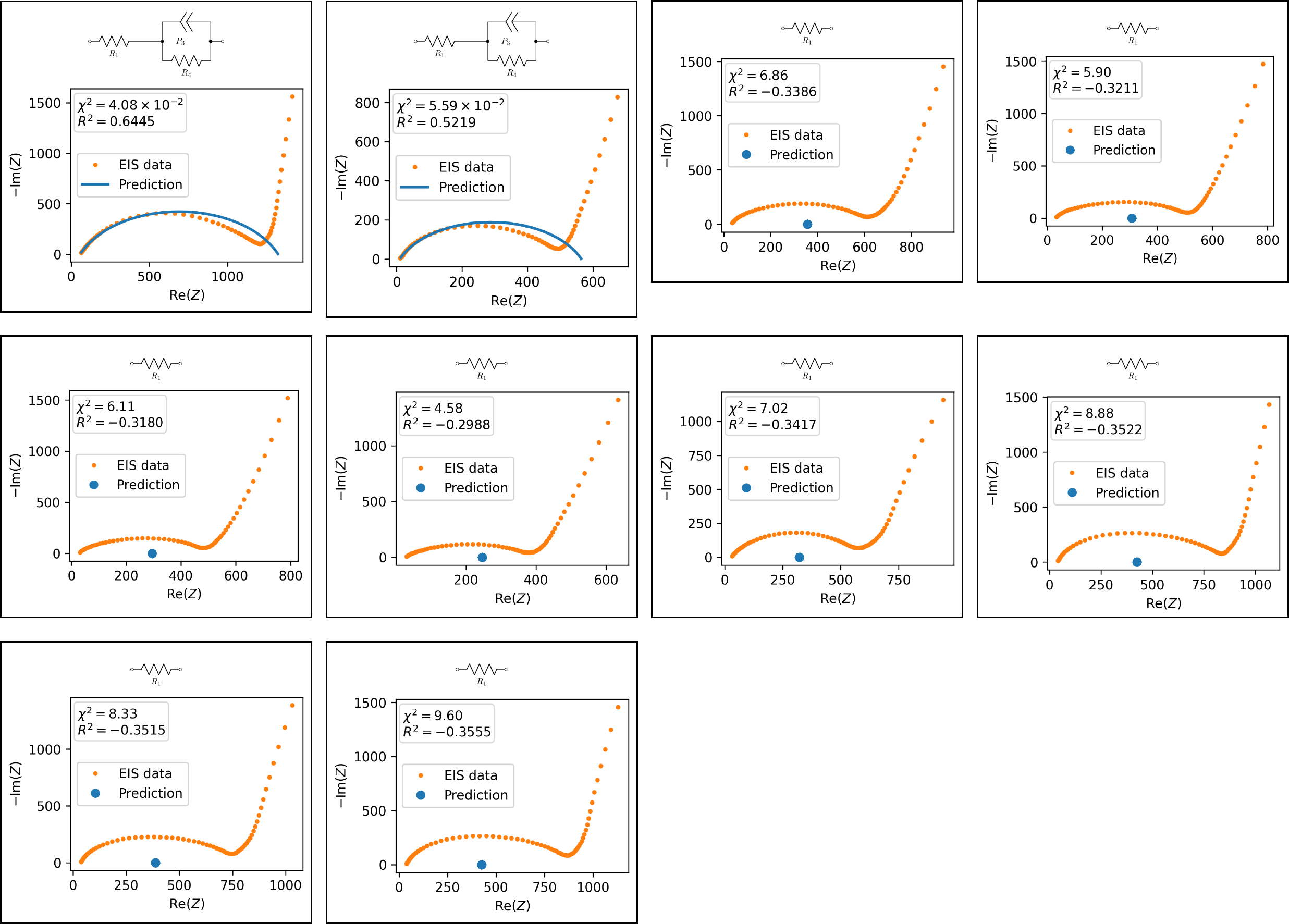}
    \caption{
        ECMs proposed by the RL agent for the battery spectra cases where the agent is unable to propose ECMs that sufficiently reproduce the measured impedance spectra with high fidelity.
        Each panel shows the proposed ECM, the corresponding predicted impedance spectrum, the measured impedance spectrum, as well as the $\chi^2$ and $R^2$ values quantifying the goodness of fit.
    }
    \label{fig:battery-failed}
\end{figure}

\begin{figure}[!hbt]
    \centering
    \includegraphics[scale=0.33]{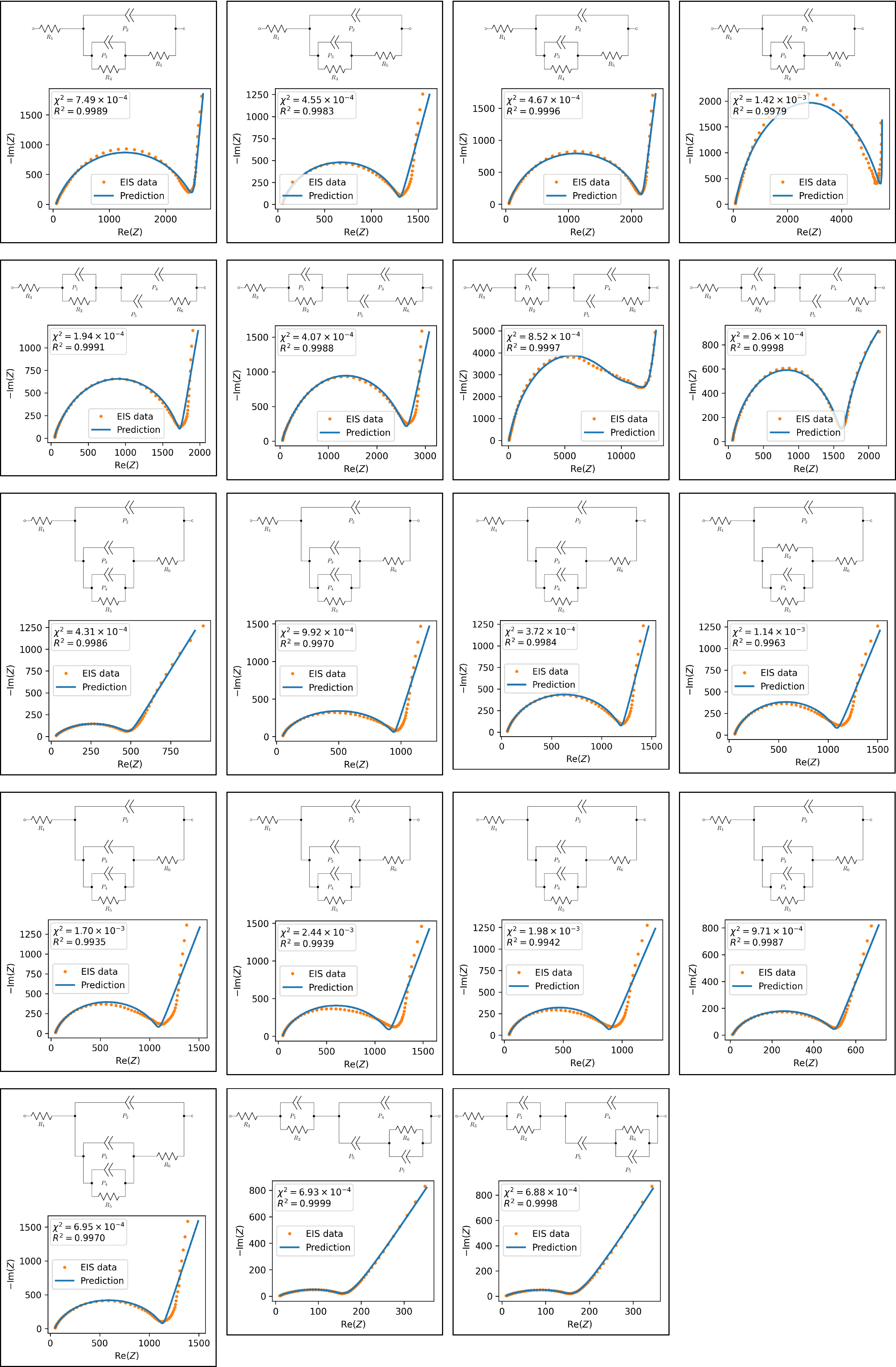}
    \caption{
        ECMs proposed by the RL agent for the battery spectra cases where the agent is able to propose ECMs that reproduce the measured impedance spectra with high fidelity.
        Each panel shows the proposed ECM, the corresponding predicted impedance spectrum, the measured impedance spectrum, as well as the $\chi^2$ and $R^2$ values quantifying the goodness of fit.
    }
    \label{fig:battery-results-success}
\end{figure}

In this section, we present the ECMs suggested by the reference agent used for demonstration on the battery data, along with the corresponding Nyquist plots comparing the impedance spectra predicted by the proposed ECMs with the measured data.
Figure~\ref{fig:battery-failed} shows the cases where the suggested ECMs are insufficient to reproduce the measured impedance spectra.
In comparison, Fig.~\ref{fig:battery-results-success} shows the cases where the reference agent proposes ECMs that reproduce the measured impedance spectra with high fidelity, as indicated by the displayed $\chi^2$ and $R^2$ values on each panel.
These cases account for $65.5\%$ of the total battery spectra.

\begin{figure}[!hbt]
    \centering
    \includegraphics[width=\textwidth]{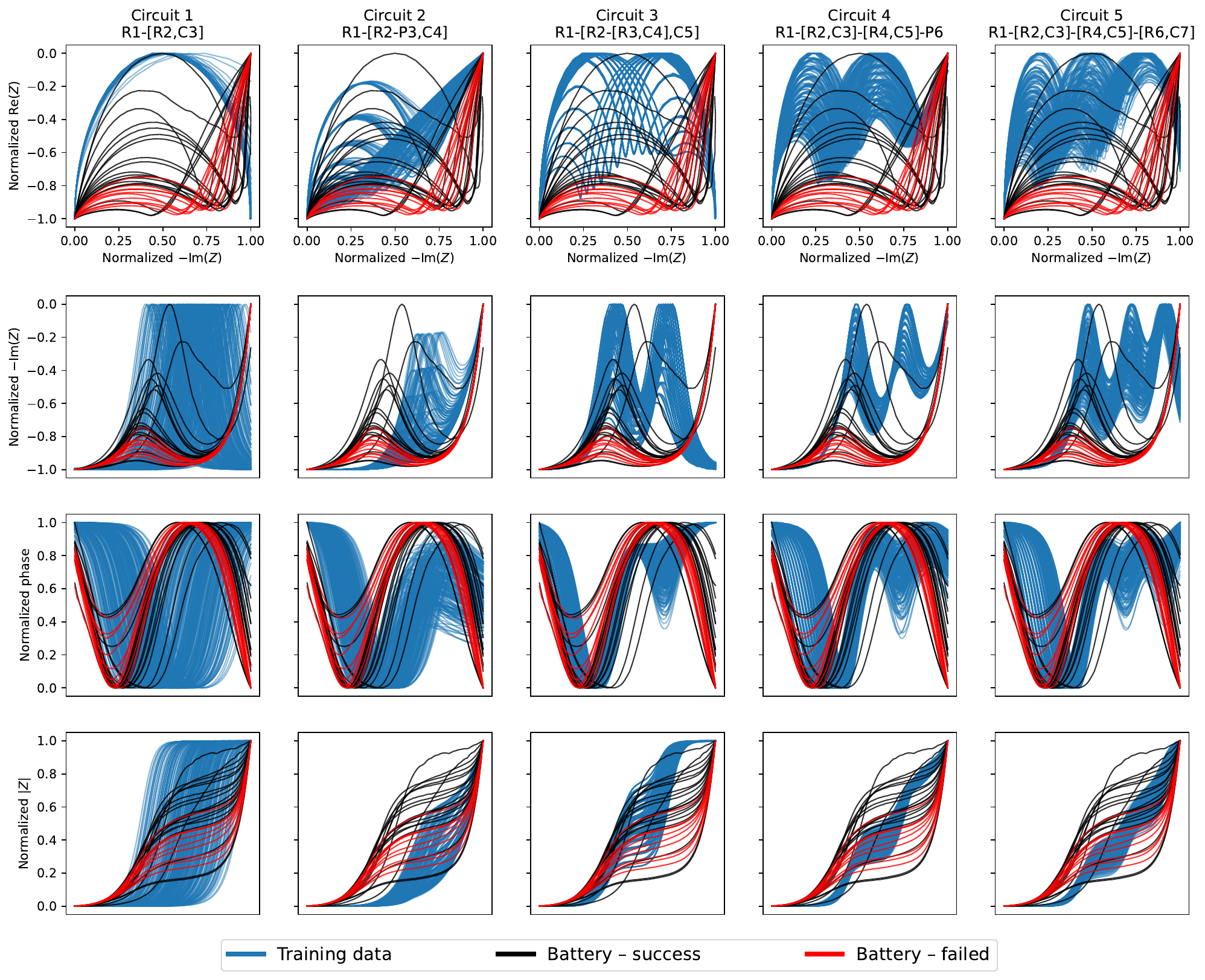}
    \caption{
        Comparison of the training data with the experimental battery spectra.
        Columns correspond to the five reference circuit families used for training.
        The rows show the normalized Nyquist representation and the impedance features supplied to the reference agent, including the imaginary impedance, phase, and magnitude.
        Blue curves denote the training spectra, while black and red curves denote battery spectra for which the reference agent was successful and failed, respectively.
    }
    \label{fig:training-battery-comparison}
\end{figure}

\begin{figure}[!hbt]
    \centering
    \includegraphics[width=0.67\textwidth]{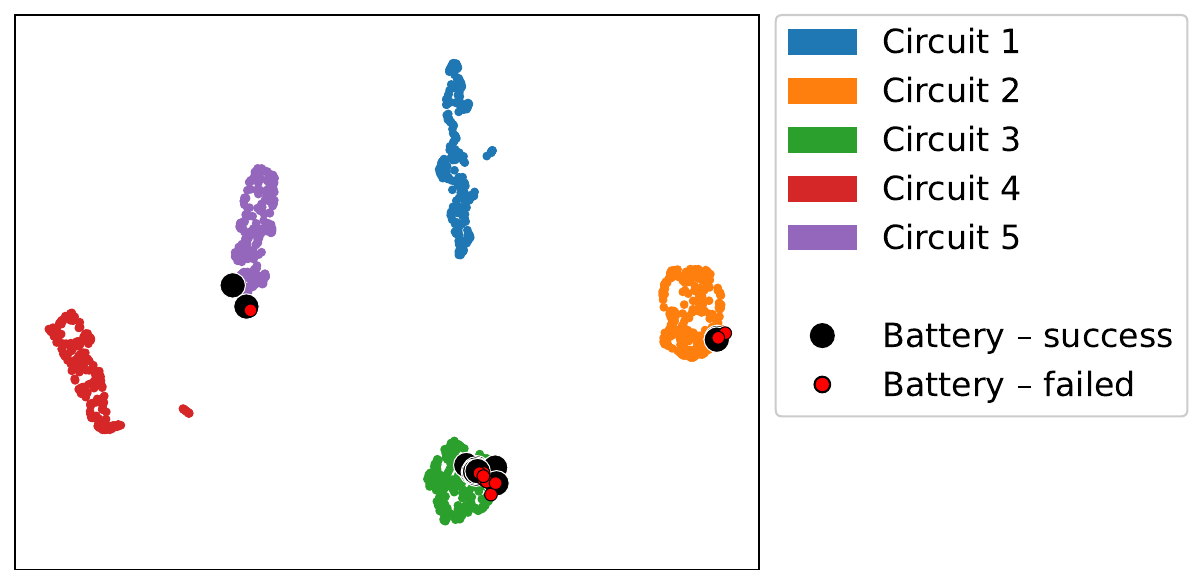}
    \caption{
        UMAP projection of the impedance representation used as input to the reference agent, comparing the training data with the experimental battery spectra.
        The five reference circuit families form distinct clusters.
        Battery spectra are shown according to the reference-agent outcome, with black markers indicating successful cases and red markers indicating failed cases.
    }
    \label{fig:training-battery-umap}
\end{figure}

Additionally, Fig.~\ref{fig:training-battery-comparison} compares the training spectra (blue curves) with the successful (black curves) and failed (red curves) battery cases using the normalized Nyquist plots and the impedance features supplied to the reference agent.
Although these plots provide a human-readable comparison, it is difficult to determine from them alone which spectral characteristics are captured well by the learned policy because the DDQN transforms the impedance input and current circuit state into an internal latent representation before selecting an action.
Spectra that appear similar visually may therefore be represented differently by the network, and vice versa.
The UMAP projection in Fig.~\ref{fig:training-battery-umap} provides a complementary view, showing that battery spectra lie near the peripheries of the distinct clusters formed by the training data.
Because the policy is learned from this training distribution, this suggests that limited training-data coverage may contribute to the variable behavior observed on the battery spectra.
A future extension could assess confidence in these predictions using uncertainty-quantification approaches such as ensembles of independently trained agents, although this is outside the scope of the present work.

\section{Comparison with AutoEIS}
\label{sec:comparison-autoeis}

To provide an external point of reference for the reference agent developed using AutoREC, we compare its results with AutoEIS, an automated ECM identification workflow that uses gene expression programming (GEP) to generate candidate circuits, followed by candidate filtering, Bayesian inference, and metrics evaluation \cite{zhang_editors_2023}.
Unlike the AutoREC reference agent, which applies a previously trained policy to generate an ECM, AutoEIS performs a new stochastic GEP search for each spectrum and may retain multiple candidate circuits.
Because of these differences, the comparison is intended to contextualize the behavior and computational characteristics of the reference implementation rather than to treat the two workflows as directly equivalent.

\subsection{Synthetic data}
\label{sec:synthetic-data}

For the synthetic-data comparison, AutoEIS was applied to the $1,350$ spectra used to train the AutoREC reference agent rather than the smaller held-out test set of $150$ spectra.
We use the larger dataset to obtain a more representative characterization of the AutoEIS outputs and computational cost, rather than as a held-out benchmark of generalization performance.
Each spectrum was evaluated once using a single GEP random seed.
Because GEP is stochastic, repeated evaluations with different random seeds could produce different candidate circuits and potentially improve the resulting fits, but would also incur additional computational cost.
The AutoEIS evaluations were perform under the same hardware and software environment as the AutoREC evaluations, with computations restricted to a single CPU core.

For each spectrum, we record whether at least one candidate circuit remains after completion of the AutoEIS workflow and define the corresponding fraction as the \emph{candidate-return rate}.
As reported in Table~\ref{tab:autoeis-success}, the candidate-return rate is $96.30\%$ for Circuit 1 and ranges from $34.81\%$ to $45.19\%$ for Circuits 2--5.
This quantity is analogous to but distinct from the AutoREC success rate, which additionally requires the generated ECM to satisfy the data-dependent $\chi^2_{\mathrm{thresh}}$ criterion.

\begin{table*}[!hbt]
    \centering
    \caption{
        Candidate-return rates of AutoEIS on the synthetic training set, reported separately for each reference topology.
    }
    \label{tab:autoeis-success}
    \setlength{\tabcolsep}{12pt}
    \begin{tabular}{lccccc}
      \toprule
      & \textbf{Circuit 1} & \textbf{Circuit 2} & \textbf{Circuit 3} & \textbf{Circuit 4} & \textbf{Circuit 5} \\
      \midrule
      \makecell[l]{Candidate-return\\rate (\%)}
      & 96.30 & 37.78 & 34.81 & 42.22 & 45.19 \\
      \bottomrule
    \end{tabular}
\end{table*}

\begin{figure}[!hbt]
    \centering
    \includegraphics[scale=0.33]{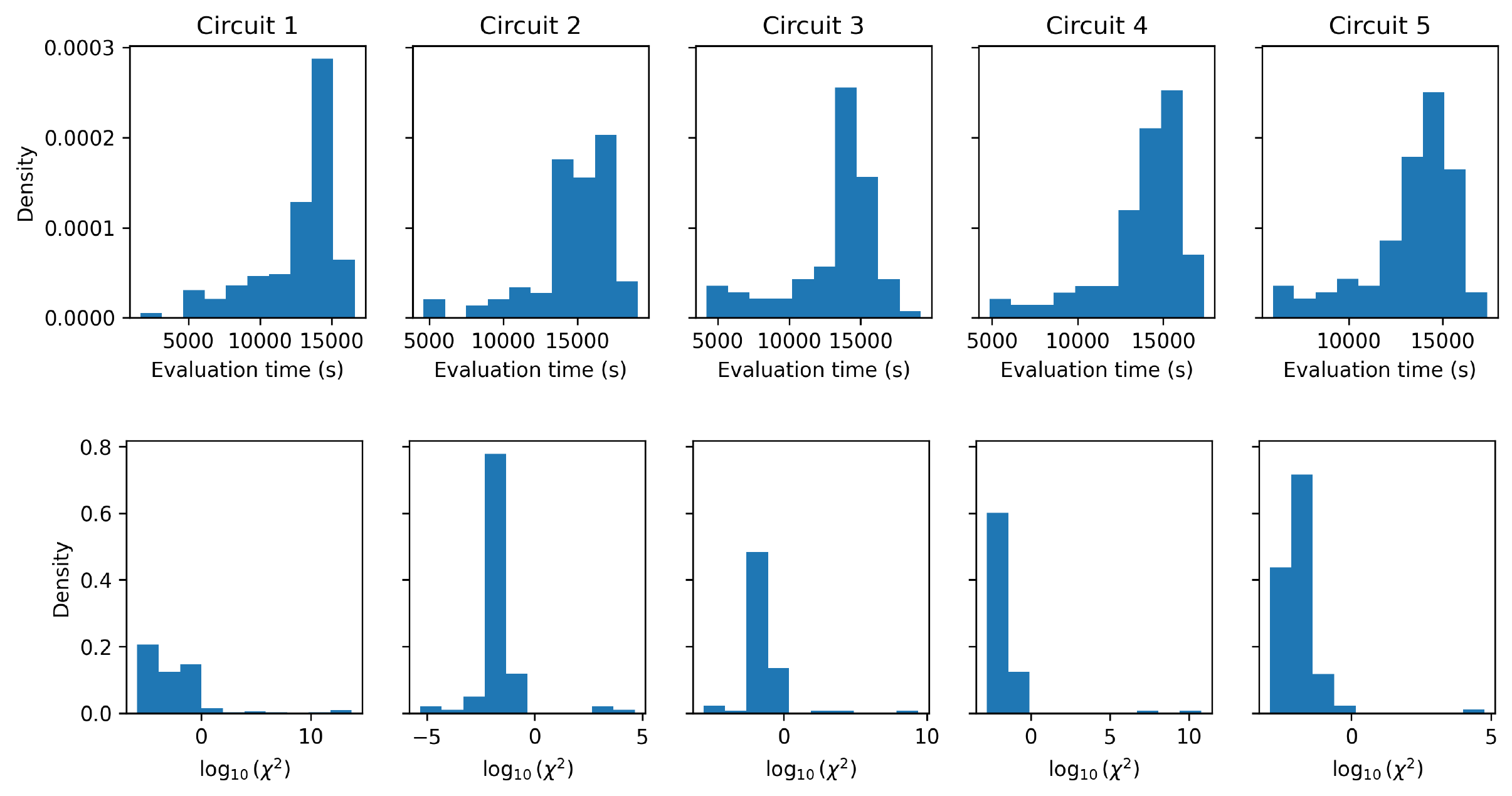}
    \caption{
        Distributions of AutoEIS evaluation time and fit quality for the synthetic training data, shown separately for each reference topology.
        The top row shows the evaluation time required to complete the AutoEIS workflow, while the bottom row shows the corresponding $\log_{10}(\chi^2)$ values.
        Only spectra for which AutoEIS retained at least one candidate circuit are included.
        When multiple candidates were retained, the circuit with the lowest WAIC value was used to calculate $\chi^2$.
    }
    \label{fig:autoeis-synthetic}
\end{figure}

When multiple candidate circuits are retained by AutoEIS, the circuit with the lowest average widely applicable information criterion (WAIC) is selected for fit-quality analysis, and its $\chi^2$ value is calculated.
For spectra with at least one retained candidate, Fig.~\ref{fig:autoeis-synthetic} shows the distributions of $\log_{10}(\chi^2)$ and evaluation time.
The $\chi^2$ distributions span several orders of magnitude, indicating that retention of a candidate by the AutoEIS workflow does not necessarily imply a uniformly high-quality fit.

The reported evaluation times cover the complete per-spectrum workflows for both methods.
For AutoEIS, this includes GEP-based circuit generation, physics-based filtering, Bayesian inference with $2,500$ warmup steps and $1,000$ inference steps, and metrics calculation.
For AutoREC, the evaluation time includes the complete circuit-generation process, from the initial ECM to the final fitted ECM, including iterative topology modifications and parameter fitting.
Under these configurations, AutoEIS evaluation times are generally on the order of $10^4$~CPU-s per spectrum (see Fig.~\ref{fig:autoeis-synthetic}, top row).
For reference, successful evaluations by the AutoREC reference agent satisfy the data-dependent $\chi^2_{\mathrm{thresh}}$ criterion, for which $\chi^2_{\mathrm{thresh}} \leq 6.23\times10^{-5}$ for the synthetic data, and in practice yield $\chi^2<1.60\times10^{-5}$.
Evaluation with the trained reference agent requires less than $33$~CPU-s per spectrum on the synthetic training data.

\subsection{Corrosion, OER and CO$_2$ reduction data}
\label{sec:experimental-data}

\begin{figure}[!htb]
    \centering
    \includegraphics[width=0.67\textwidth]{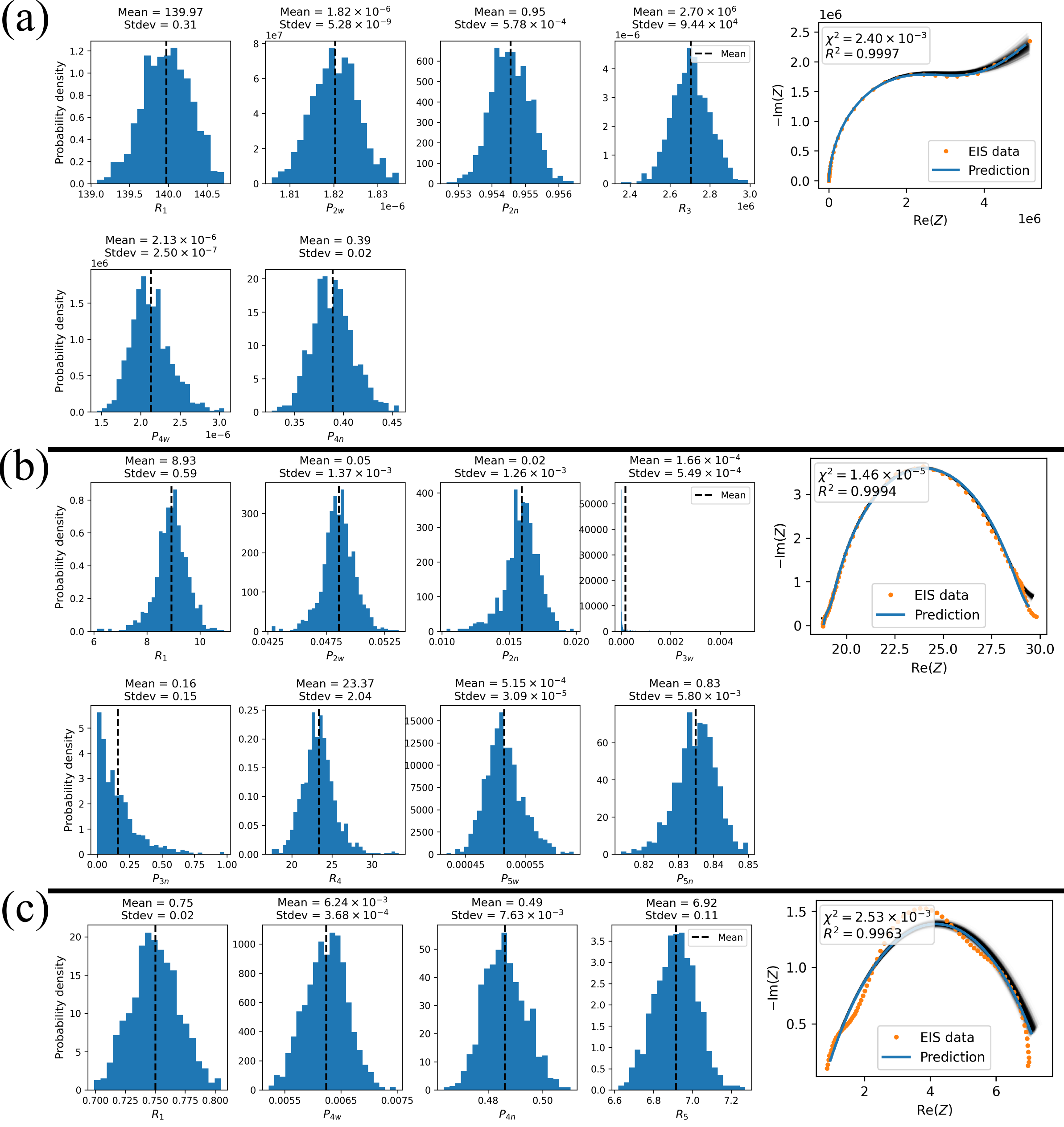}
    \caption{
        Posterior distributions of the model parameters and corresponding posterior-predictive impedance responses for (a) corrosion, (b) OER, and (c) CO$_2$ reduction data.
        The histograms show the marginal posterior distributions of the model parameters, with dashed vertical lines indicating their posterior means.
        In the Nyquist plots, the experimental EIS data are shown as orange markers, while the blue curve shows the fitted response of the ECM suggested by the reference agent.
        The black curves show the distribution of impedance responses obtained by evaluating the ECM for the individual posterior samples.
    }
    \label{fig:experimental-bi}
\end{figure}

For the corrosion, OER, and CO$_2$ reduction data, AutoEIS results has been reported previously by \citet{zhang_editors_2023}, and the qualitative comparison with the reference agent is discussed in the main manuscript.
Here, we further assess the ECMs suggested by the reference agent by examining the posterior distributions of their parameters, as shown in Fig.~\ref{fig:experimental-bi}.
For the corrosion and CO$_2$ reduction systems, as well as most parameters for OER, the posterior distributions are relatively narrow and approximately Gaussian, indicating that these parameters are well constrained by the experimental data within the proposed ECMs.
This does not establish that the ECM itself is uniquely appropriate, but it indicates that the data provide information that constrains these parameters.
In contrast, some parameters for the OER system exhibit broader and skewed posterior distributions, suggesting weaker parameter constraint and possible practical identifiability issues.
Despite these differences in parameter uncertainty, the posterior-predictive impedance responses remain tightly distributed for all three systems.

\subsection{Battery data}
\label{sec:battery-data}

\begin{figure}[!hbt]
    \centering
    \includegraphics[scale=0.33]{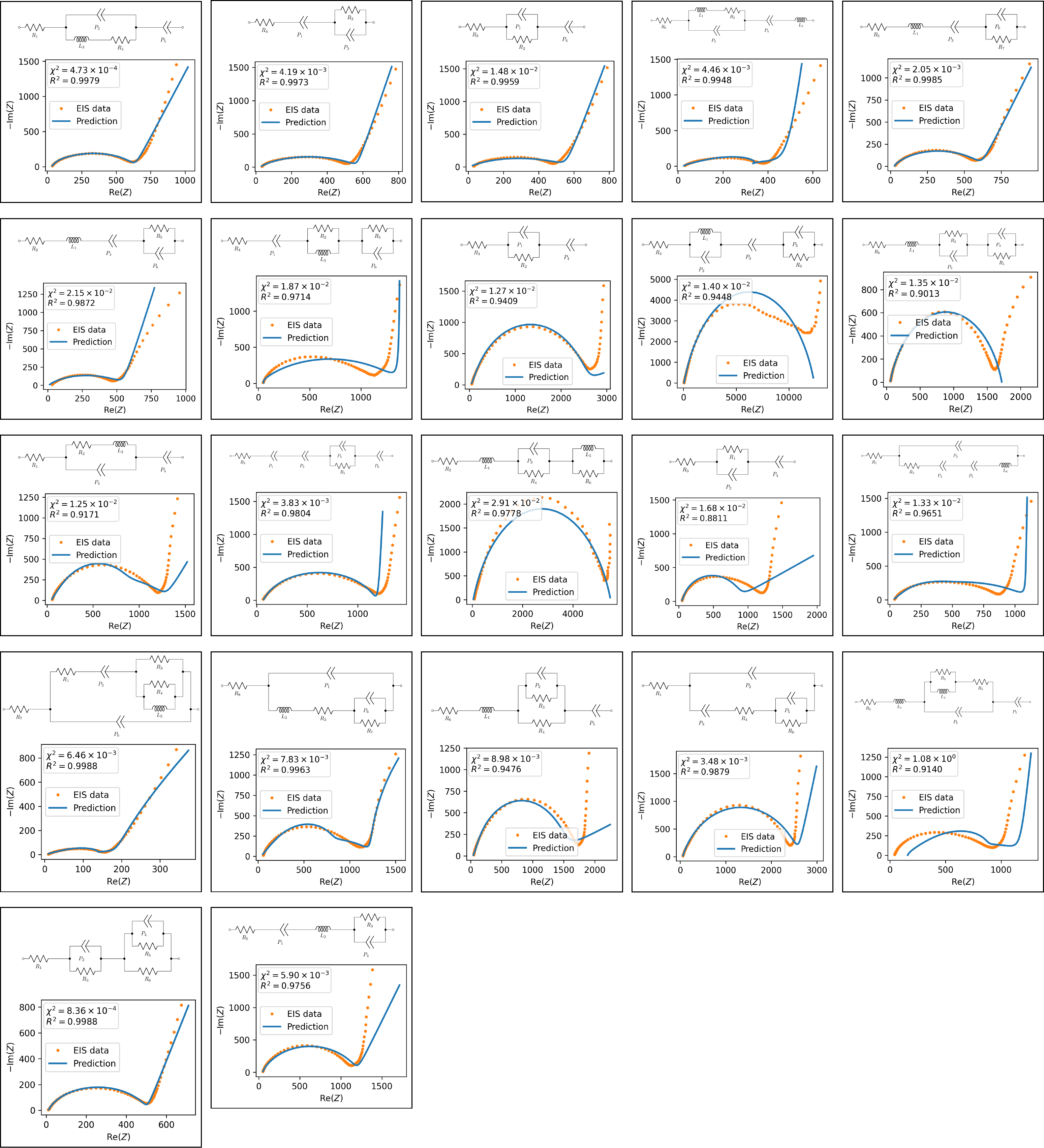}
    \caption{
        AutoEIS results for the battery EIS spectra for which at least one candidate circuit was returned.
        For each result shown, the displayed ECM is the candidate with the lowest WAIC value.
        For each spectrum, the selected ECM is shown above the corresponding Nyquist plot, comparing the experimental data (orange markers) with the impedance response predicted by the selected ECM (blue curve).
        The corresponding $\chi^2$ and $R^2$ values are reported in each panel.
    }
    \label{fig:autoeis-battery}
\end{figure}

Finally, AutoEIS returned at least one candidate circuit for 22 of the 29 battery spectra, corresponding to a candidate-return rate of $75.86\%$.
Figure~\ref{fig:autoeis-battery} shows the selected ECM and corresponding fitted and measured Nyquist spectra for these 22 cases, together with their $\chi^2$ and $R^2$ values.
As in Sec.~\ref{sec:synthetic-data}, when multiple candidates were returned, we selected the circuit with the lowest WAIC value for presentation.
Candidate return, however, does not necessarily imply a high-quality fit as many of them have $\chi^2>10^{-2}$ and $R^2<0.95$.
However, these results are based on a single AutoEIS run for each spectrum.
Because the GEP search is stochastic, repeated runs with different random seeds may return different, and potentially better-fitting, candidates, but at additional computational cost.

\bibliographystyle{rsc}
\bibliography{references.bib, references_zotero.bib}